\documentclass[a4paper,pdflatex]{article}

\usepackage[a4paper,total={170mm,257mm},left=20mm,top=20mm]{geometry}

\usepackage{amssymb}
\usepackage{amsthm}

\usepackage{lineno}
\usepackage{authblk}

\usepackage{subfig}
\usepackage{tikz}
\usetikzlibrary{calc}
\usetikzlibrary{arrows.meta}
\usepackage{pgfplots}

\usepackage{booktabs}
\usepackage{color, colortbl}

\usepackage{algorithm}
\usepackage{algpseudocode}
\MakeRobust{\Call}

\usetikzlibrary{positioning,arrows,automata}

\usepackage[toc]{appendix}

\usepackage{pgfgantt}

\definecolor{umBlue}{RGB}{8,28,59}
\definecolor{aucosRed}{RGB}{166,25,46}
\definecolor{bgCol}{RGB}{250,250,250}

\begin{document}

\title{Monte-Carlo Tree-Search for Leveraging Performance \\ of Blackbox Job-Shop Scheduling Heuristics}

%\author{Florian Wimmenauer \and Mat\'{u}\v{s} Mihal\'{a}k \and Mark H.M. Winands}
\author[1,2]{Florian Wimmenauer}
\author[2]{Matúš Mihalák}
\author[2]{Mark H. M. Winands}
\affil[1]{Aucos, Aachen, Germany}
\affil[2]{Dept. of Advanced Computing Sciences, Maastricht University, Maastricht, the Netherlands}
\date{December 2022}

\maketitle

\begin{abstract}
In manufacturing, the production is often done on out-of-the-shelf manufacturing lines, whose underlying scheduling heuristics for the underlying scheduling problems are not known due to the intellectual property of the providers of the manufacturing lines. %Yet, for complex underlying scheduling problems, the ordering of jobs that enter the manufacturing line (and thus the underlying heuristic) can substantially influence the overall processing time, i.e., the makespan.

We consider such a setting with a black-box job-shop system, and an unknown scheduling heuristic that, for a given permutation of jobs, schedules the jobs for the black-box job-shop with the goal of minimizing the makespan. 
Here, the jobs need to enter the job-shop in the given order of the permutation, but may take different paths within the job shop, which depends on the black-box heuristic. The performance of the black-box heuristic depends on the order of the jobs, and the natural problem for the manufacturer is to find an optimum ordering of the jobs.

Facing a real-world scenario as described above, we engineer the Monte-Carlo tree-search for finding a close-to-optimum ordering of jobs. To cope with a large solutions-space in planning scenarios, a hierarchical Monte-Carlo tree search (H-MCTS) is proposed based on abstraction of jobs.
On synthetic and real-life problems, H-MCTS with integrated abstraction significantly outperforms pure heuristic-based techniques as well as other Monte-Carlo search variants.

We furthermore show that, by modifying the evaluation metric in H-MCTS, it is possible to achieve other optimization objectives than what the scheduling heuristics are designed for -- e.g., minimizing the total completion time instead of the makespan. 

Our experimental observations have been also validated in real-life cases, and our H-MCTS approach has been implemented in a production plant's controller. 
\end{abstract}

%% \linenumbers

%%%%%%%%%%%%%%%%%%%%%%%%%%%%%%%%%%%%%%%%%%%%%%%%%%%%%%%%%%%%%%%%%%%%%%%%%%%%%%%%%
%%%   Introduction
%%%%%%%%%%%%%%%%%%%%%%%%%%%%%%%%%%%%%%%%%%%%%%%%%%%%%%%%%%%%%%%%%%%%%%%%%%%%%%%%%

\section{Introduction} \label{sec:intro}

Manufacturing plants often buy their manufacturing lines off-the shelf from an independent third-party. A product that is produced on the line goes through several production steps. These steps are performed at stations of the manufacturing line. Manufacturing lines consist of several stations, some of which are equivalent in their function. Complex manufacturing lines can produce several different products that differ, from the manufacturing-line point of view, in the path through the stations of the line, and in the (minimum) time needed at the stations for the respective production step.

As a result, these manufacturing lines perform non-trivial scheduling which controls the timing and routing of products through the manufacturing lines.
Often, these scheduling heuristics is encapsulated in the production-line controller, and are not know nor accessible for altering to the purchaser of the manufacturing line. 
There can be many reasons for this; the typical ones are the intellectual property, or safety regulations of the manufacturing line (which results in an outside-access restriction).

Thus, the underlying scheduling heuristic of the manufacturing line is seen as a black box to the planners of the manufacture, which need to decide 
%what products and in which ... they enter ...
the \emph{plan}, i.e., an ordering, in which the products, also called \emph{jobs}, enter the black-box manufacturing line. Often, the goal of the planners is to minimize the makespan -- the time when the last job was processed (produced) on the manufacturing line. 

A straightforward approach to finding an optimal solution (a plan) is to enumerate all possible solutions (i.e., all permutations of jobs) and keep the solution with the smallest makespan.
Such an enumeration can be represented by a so-called \emph{search tree} -- a rooted tree, where the root is an empty plan and each child node is created by adding one job to the first available position of the partial plan represented by its parent. Thus, the number of children of a node is equal to the number of jobs that do not appear in the partial plan of the node.
Each leaf node of the tree represents a complete plan. Thus, if $n$ is the number of jobs, there are $n!$ many leaves in the tree.
A recursive way to enumerate all plans is equivalent to a depth-first search of the search tree.
In large industrial manufacturing lines, the number of jobs that need to be planned can be up to thousands. To cope with the exponentially large solution space, a more sophisticated approach than a straightforward brute-force enumeration is needed.

A common method for approaching such a combinatorial optimisation problem is branch-and-bound \cite{little1963algorithm}, where upper and lower bounds on the makespan of the leaves of branches in the tree are computed and used for pruning to reduce the search space.
Non-trivial lower and upper bounds on the makespan can often be derived when the entire setting of the underlying scheduling problem (e.g., the processing times at the stations on the manufacturing lines) and the scheduling heuristic is known.
However, in our case, as the underlying heuristic is a blackbox, no upper bound on the schedule can be formulated, which renders standard branch-and-bound methods not applicable.

To cope with the lack of knowledge, we consider a stochastic (sampling-based) variant of the branch-and-bound search -- a Monte-Carlo tree search -- and engineer a non-trivial version of the method to find near-to-optimum plans.
To apply any kind of Monte-Carlo tree-search, we need good estimates of the overall processing times of products on the black-box production line. In general, these processing times may be known, or can be learned from time logs of past production, or from simulation tools that are often delivered with the manufacturing line. In our use-case, we successfully apply neural networks to learn these from past sample data.
We observe that standard Monte-Carlo tree-search methods fail, and, based on the experimental evaluations, we engineer a \emph{hierarchical Monte-Carlo tree-search} \cite{vien2015hierarchical, bai2016markovian} method based on the \emph{abstraction of jobs}.
More specifically, we group similar jobs into one abstract job type.
A key novelty of this work is that the construction of abstraction hierarchy is integrated into the search operations and therefore automated.
We observe that our approach is scalable regarding the number of jobs. 
Together with the modelling of the black-box scheduling heuristic by neural networks, the proposed approach is thus generically applicable to different production plants, independent of the black-box scheduling heurstics.

%
%With the heuristic seen as a blackbox, the schedule optimisation task is transformed to finding a permutation of jobs that minimises the total makespan.
%Figure \ref{fig:approach} visualises this scenario, where the blackbox scheduling heuristic takes a planned job sequence as input and outputs the completion times of each of these jobs. 
%
%Before being scheduled, the jobs can request resources such as containers, which will return at the job’s completion.
%
%Noteworthy is that, in this case, the concepts of planning and scheduling need to be distinguished. 
%
%A plan is the sequence of jobs provided to the scheduler, whereas a schedule is what the scheduler outputs based on the given plan.

%\begin{figure}
%	\centering
%	\input{Chapter1/figures/approach}
%	\captionsetup{width=.9\linewidth}
%	\caption{A planned job sequence consumes external resources and is fed into the scheduler, which outputs the job's completion times. Upon job completion, resources are freed for reuse. The blue components are controlled by the planning system and the black ones by the blackbox scheduling heuristic.}%
%	\label{fig:approach}%
%\end{figure}

%This article seeks to leverage the performance of existing JSP heuristics on large-scale industrial problems by planning the jobs provided to the scheduler.

%----------------outlines other parts-------------
\paragraph{Outline of the paper.}
The remainder of the paper is organized as follows. First, Section~\ref{sec:related work} discusses the related concepts and previous work in planning using Monte-Carlo based search.
Section~\ref{sec:problem definition} formally defines the elementary planning problem, and also defines a more-refined variant of the problem with resources, which reflects the details of the real-world use-case that we experiment with.
Next, Section~\ref{sec:mcts} describes in detail the proposed approach to enhancing JSP heuristic performance with Monte-Carlo Tree Search. 
Subsequently, Section~\ref{sec:experimental setup} presents the experimental set-up. Section~\ref{sec:results} discusses the outcomes and discussions of the experimental results.
Finally, Section~\ref{sec:conclusions} concludes the article by summarizing the key contributions and discussing directions of further research.

%%%%%%%%%%%%%%%%%%%%%%%%%%%%%%%%%%%%%%%%%%%%%%%%%%%%%%%%%%%%%%%%%%%%%%%%%%%%%%%%%
%%%   Related Work
%%%%%%%%%%%%%%%%%%%%%%%%%%%%%%%%%%%%%%%%%%%%%%%%%%%%%%%%%%%%%%%%%%%%%%%%%%%%%%%%%

\section{Related Work} \label{sec:related work}

Monte-Carlo tree-search has first been developed for algorithms to play board games \cite{coulom2006efficient}. In that context, the vertices of the search tree represent the states of the game and the edges represent the moves of the corresponding players.
Since then, there has been a substantial research on variants of Monte-Carlo tree-search methods, mainly fueled by the interest in game play.
Guiding the tree search with \emph{upper confidence bound applied to trees} (UCT) \cite{kocsis2006bandit} is one of the most successful techniques for agents in two-player games \cite{gelly2007combining, finnsson2010learning}. Notably, MCTS technique, combined with neural networks, has achieved tremendous success for the board game Go \cite{gelly2011monte,silver2016mastering}.

Techniques inspired by Monte-Carlo tree-search (MCTS) have gained increasing interest in various research fields beyond playing board games.
%
% Is there still more to write about?
%
Besides in two-player games, MCTS variants for one-player games \cite{mehat2010combining, schadd2012single} applied in the field of combinatorial optimization \cite{Coutoux11b,rossi2019} and planning \cite{zhu2014,hennes2015} have shown promising results. Recently, \cite{segler2017learning} have proposed a planner based on MCTS and deep-learning for chemical synthesises which achieved human-level performance in generating retro-synthetic routes. 
It has been observed that one particular strength of MCTS-based methods have minimal need of problem-specific domain knowledge \cite{lubosch2018industrial}, which makes the approach applicable to a diverse group of optimization and planning problems.
We use this property in our research as well, and apply MCTS to our setting with a black-box scheduling heuristic.

The research interest in single-player games and optimization led to the development of specialised techniques such as \emph{nested Monte-Carlo search} (NMCS) \cite{cazenave2009nested} and \emph{nested rollout policy adaptation} (NRPA) \cite{rosin2011nested}, which outperform standard approaches by using nested levels to guide the search more efficiently \cite{cazenave2009monte}. 
Most Monte-Carlo tree-search approaches for planning deal with a moderate number of jobs, where the processing of jobs is stochastic and known \cite{van2013guided, simroth2015job, runarsson2012pilot}. 
More recently, \emph{hierarchical MCTS} has been proposed \cite{vien2015hierarchical, bai2016markovian}, addressing computational-time problems caused by exponential search-tree growth when search depth increases.

%%%%%%%%%%%%%%%%%%%%%%%%%%%%%%%%%%%%%%%%%%%%%%%%%%%%%%%%%%%%%%%%%%%%%%%%%%%%%%%%%
%%%   Problem Definition
%%%%%%%%%%%%%%%%%%%%%%%%%%%%%%%%%%%%%%%%%%%%%%%%%%%%%%%%%%%%%%%%%%%%%%%%%%%%%%%%%

\section{Terminology and Problem Definition} 
\label{sec:problem definition}

In scheduling, a \emph{job} is a work/production item that requires for its completion one or more resources over certain period of time. For example, such a resource can be a worker or a machine. In scheduling, any such resource is referred to as a \emph{machine}. The time required for every job to be spend at certain machine is called the \emph{processing time} of that job on that machine.

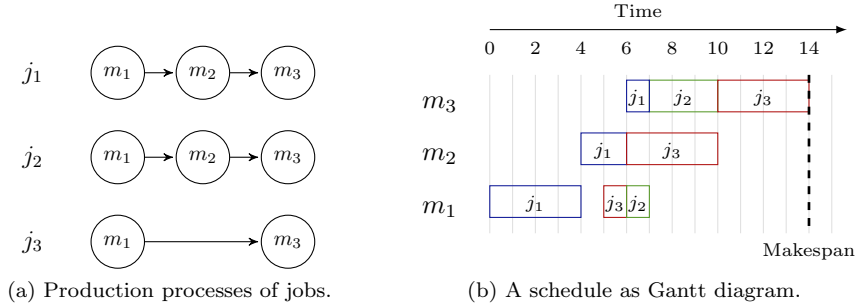
\begin{figure}%
	\centering
	
	\subfloat[Production processes of jobs.]{\begin{tikzpicture}[->,>=stealth',shorten >=1pt,auto,node distance=1.4cm]
	\tikzstyle{machine}=[fill=none,draw=black,text=black]
	\tikzstyle{job}=[fill=none,draw=none,text=black]
	\node[state, scale=.8][job] (j1) {\large$j_1$};
	\node[state, scale=.8][job] (j2) [below of=j1] {\large$j_2$};
	\node[state, scale=.8][job] (j3) [below of=j2] {\large$j_3$};
	\node[state, scale=.8]      (j1m1) [right of=j1]   {$m_1$};
	\node[state, scale=.8]      (j1m2) [right of=j1m1] {$m_2$};
	\node[state, scale=.8]      (j1m3) [right of=j1m2] {$m_3$};
	\node[state, scale=.8]      (j2m1) [below of=j1m1] {$m_1$};
	\node[state, scale=.8]      (j2m2) [below of=j1m2] {$m_2$};
	\node[state, scale=.8]      (j2m3) [below of=j1m3] {$m_3$};
	\node[state, scale=.8]      (j3m1) [below of=j2m1] {$m_1$};
	%\node[state, scale=.8] (j3m2) [below of=j2m2] {$m_1$};
	\node[state, scale=.8]      (j3m3) [below of=j2m3] {$m_3$};
	\path (j1m1) edge (j1m2) (j1m2) edge (j1m3);
	\path (j2m1) edge (j2m2) (j2m2) edge (j2m3);
	\path (j3m1) edge (j3m3);
%	
%	\path[dashed] (j1m1) edge (j2m1) (j2m1) edge (j3m1);
%	\path[dashed] (j1m2) edge (j2m2);
%	\path[dashed] (j1m3) edge[bend left] (j3m3) (j3m3) edge (j2m3);
\end{tikzpicture}\textsl{} \label{fig:jsp_a}}%
	\qquad
	\subfloat[A schedule as Gantt diagram.]{\definecolor{blue}{RGB}{17,27,152}
\definecolor{red}{RGB}{176,19,18}
\definecolor{green}{RGB}{89,154,43}

\begin{tikzpicture}
\begin{ganttchart}[
	x unit=0.3cm,
	y unit chart=0.7cm,
	canvas/.style={draw=none,fill=none}, % remove canvas borders, etc
	vgrid={*1{draw=black!12}},           % vertical gray lines every unit
	inline,                              % draw bars inline
	group/.style={draw=none,fill=none},  % remove group borders, etc
	bar top shift=0.1,                   % give bar 10% padding top/bottom
	bar height=0.6,                      % bar size 70% of vertical space
	y unit title=0.5cm,                  % crop titles a little smaller
	title/.style={draw=none,fill=none},  % remove title borders, etc
	include title in canvas=false        % no vertical grid in title
	]{-1}{15}
	\gantttitle{\scriptsize Time}{15}\\
	\gantttitle{\scriptsize 0}{2}
	\gantttitle{\scriptsize 2}{2}
	\gantttitle{\scriptsize 4}{2}
	\gantttitle{\scriptsize 6}{2}
	\gantttitle{\scriptsize 8}{2}
	\gantttitle{\scriptsize 10}{2}
	\gantttitle{\scriptsize 12}{2} 
	\gantttitle{\scriptsize 14}{2} \\
	%\gantttitle{16}{2} \\
	
	\ganttgroup[inline=false]{$m_3$}{0}{1}
	\ganttbar[name=m1j2, bar/.style={draw=green}]{\scriptsize$j_2$}{7}{9}
	\ganttbar[name=m1j1, bar/.style={draw=blue}]{\scriptsize$j_1$}{6}{6} 
	\ganttbar[name=m1j3, bar/.style={draw=red}]{\scriptsize$j_3$}{10}{13} \\

	\ganttgroup[inline=false]{$m_2$}{0}{1}
	\ganttbar[name=m2j1, bar/.style={draw=blue}]{\scriptsize$j_1$}{4}{5}
	\ganttbar[name=m2j3, bar/.style={draw=red}]{\scriptsize$j_3$}{6}{9} \\
	
	\ganttgroup[inline=false]{$m_1$}{0}{1}
	\ganttvrule{\scriptsize Makespan}{13}
	\ganttbar[name=m3j1, bar/.style={draw=blue}]{\scriptsize$j_1$}{0}{3}
	\ganttbar[name=m3j3, bar/.style={draw=red}]{\scriptsize$j_3$}{5}{5}
	\ganttbar[name=m3j2, bar/.style={draw=green}]{\scriptsize$j_2$}{6}{6}
	%\ganttlink[link type=straight]{m3j1}{m2j1}
	
\end{ganttchart}
\draw [-latex] (0.3,-0.4)  -- (5.0,-0.4) ;
%\node[draw, rotate=90, draw=none] at (-1,-2) {\small Jobs};

\end{tikzpicture}  \label{fig:jsp_b}}%
	\captionsetup{width=.9\linewidth}
	\caption{A job-shop with jobs $j_1,j_2,j_3$ and machines $m_1,m_2,m_3$. Subfigure (a) depicts the production processes of the jobs as directed paths. Subfigure (b) depicts a feasible schedule where every box with label $j_i$, $i=1,2,3$, corresponds to a time interval assigned to job $j_i$.}%
	\label{fig:jsp}%
\end{figure}

A \emph{job-shop} refers to a set-up where every job is processed in a sequential order by a subset of machines and the processing sequence is job-dependent.
The sequence of machines required by a job is referred to as the \emph{production process} of this job.
A \emph{schedule} for a job-shop setting is an assignment that assigns for every job and every machine of the production process of the job a time interval of length equal to the required processing time of that job on that machine, and such that all intervals from any machine do not overlap, all intervals assigned to any job do not overlap, and all intervals assigned to any job adhere with its chronological order to the processing sequence of the job in question.
Figure \ref{fig:jsp} shows an example of a schedule of a small job-shop setting with three jobs and three machines.

For a schedule of a job-shop setting, the \emph{starting time} of a job $j$ is the start of the first interval assigned to job $j$ in the schedule. Similarly, the \emph{completion time} of a job $j$ is the end of the last interval assigned to job $j$ in the schedule.
The makespan of a schedule is the largest completion time of the jobs.
The \emph{deadline} or \emph{due date} of a job is a desired latest completion time of the job.
The \emph{lateness} of a job is the difference between the completion time and the deadline of a job. Observe that lateness can be negative, if the job is completed before its deadline.
The \emph{tardiness} of a job is the maximum of two values: the lateness and zero.

We define the \emph{job-shop problem}, abbreviated as JSP, as the optimization problem of finding a schedule of minimum makespan.

We consider a specific class of algorithms for JSP, which resembles online algorithms: for a given permutation of the jobs, the algorithm starts scheduling the jobs on the machines in the given order by the permutation, and assigns the starting time of the considered job (i.e., the start of the first interval assigned to the job) as soon as possible. The algorithm schedules the remaining intervals of the jobs with the current knowledge of the jobs that were considered so far, i.e., the algorithm does not take the existence of future jobs into account.

Thus, the above algorithm can be decomposed into two: a \emph{sequencing algorithm} that computes a permutation of the job, and a \emph{black-box algorithm} that assigns intervals to jobs in the order given by the compute permutation, adhering to the rules explained above.

In view of the above definition, we call the computed permutation a \emph{plan}, and the assignment of intervals a \emph{schedule}. 

The optimization problem that we study in this paper is the following: given a set $J$ of $n$ jobs, without the knowledge of the processing times, of the production processes, or the number of machines, and given the black-box algorithm as an \emph{oracle} that for any given permutation of jobs returns the start and completion times of the jobs, find an optimum permutation that, together with the black-box algorithm, minimizes the makespan.

In our experimental evaulation, we will consider a real-world use case, which is a slightly generalized job-shop setting. In particular, in that use-case, there are jobs and a limited number of \emph{resources} of few types, and every job needs to enter the job-shop together with a resource of a specific type. After the job is completed, the associated resource will become available for usage for different jobs. We will discuss in details the set-up of this variant of job-shop later in Section~\ref{subsec:industrial_case}.

%%%%%%%%%%%%%%%%%%%%%%%%%%%%%%%%%%%%%%%%%%%%%%%%%%%%%%%%%%%%%%%%%%%%%%%%%%%%%%%%%
%%%   MCTS for Black-Box Job-Shop Scheduling
%%%%%%%%%%%%%%%%%%%%%%%%%%%%%%%%%%%%%%%%%%%%%%%%%%%%%%%%%%%%%%%%%%%%%%%%%%%%%%%%%

\section{Monte-Carlo Tree Search for Blackbox Job-Shop Scheduling}
\label{sec:mcts}

In this section we explain in details the tweaking and engineering of our search approach for a good plan in the search tree. Our approach is a non-trivial variant of the plain Monte-Carlo tree-search (MCTS). We first discuss the plain MCTS in Subsection \ref{subsec:MCTS}. 
Next, several enhancements for MCTS are discussed in Subsection \ref{subsec:enhancements}. The MCTS variant to deal with the combinatorial nature of the search problem, called Hierarchical Monte-Carlo Tree Search (H-MCTS), is introduced in Subsection \ref{subsec:HMCTS}. Its abstraction mechanism is further detailed in Subsection \ref{subsec:jobAbstraction}. The modelling of the underlying scheduling heuristic for (H-)MCTS is discussed in Subsection \ref{sec:learning}. Finally, it is discussed how a generated plan by a simulation is being assessed in Subsection \ref{subsec:TSE}.

\subsection{Monte-Carlo Tree Search}\label{subsec:MCTS}
Monte-Carlo Tree Search (MCTS) \cite{coulom2006efficient} is best-first search technique guided by Monte-Carlo evaluations. Initially, the search starts with an empty plan as the root node. From there a tree is gradually grown by executing the selection, roll-out, expansion, and backpropagation phases. Such an iteration is called a full simulation. 

In the first phase, starting at the root node, the tree is traversed using a \textit{selection} strategy such as UCT \cite{kocsis2006bandit}. With UCT, at each node $p$ a child $i$ maximising the following equation is selected:
\begin{equation}
	v_i=s_i + C  \sqrt{\frac{\ln(n_p)}{n_i}}
	\label{eq:uct}
\end{equation} 
Here  $s_i = \bar{s_i} \in [0,1]$ is normalised average score, $n_p$ is the number of visits of $i$'s parent node and $n_i$ is the number of visits of a child node $i$.  UCT balances the exploration of under-explored nodes and the exploitation of nodes with known good values.
The denominator of the second term in Equation \ref{eq:uct} ensures that rarely visited child nodes receive higher scores and are eventually investigated.

\begin{figure}[t]%
	\centering
	%\qquad
	\begin{tikzpicture}[
scale=0.7, every node/.style={scale=0.7},
abstract/.style={rectangle, draw=blue, rounded corners=0.5mm, fill=none,
	text centered, anchor=north, text=black, minimum size=5mm},
abstractArc/.style={rectangle, draw=blue, rounded corners=0.5mm, fill=none,
	text centered, anchor=north, text=black, minimum size=5mm},
root/.style={circle, draw=blue, fill=none,
	text centered, anchor=north, text=black},
primitive/.style={circle, draw=red, fill=none,
	text centered, anchor=north, text=black, minimum size=5mm},
primitiveArc/.style={circle, draw=red, fill=none,
	text centered, anchor=north, text=black, minimum size=5mm},
contArc/.style={dashed, draw=black, fill=none,
	text centered, anchor=north, text=black, minimum size=5mm},
abstractContArc/.style={dashed, draw=blue, fill=none,
	text centered, anchor=north, text=black, minimum size=5mm},
primitiveContArc/.style={dashed, draw=red, fill=none,
	text centered, anchor=north, text=black, minimum size=5mm},
level distance=1cm, growth parent anchor=south
]
\node (root) [abstract] {root} [->]{ 
	[sibling distance=3.5cm]
		child[primitiveArc]{ 
			[sibling distance=1cm]
			node (a) [primitive]{}{
				child[primitiveArc]{ 
					[sibling distance=4cm]
					node (ab) [primitive]{}{
						[sibling distance=0.3cm]
						child[primitiveContArc]
					}
					edge from parent node[left]{$b$}
				}
				child[primitiveArc]{ 
					[sibling distance=4cm]
					node (ac) [primitive]{}{
						[sibling distance=0.3cm]
						child[primitiveContArc]
					}
					edge from parent node[left]{$c$}
				}
				child[primitiveArc]{ 
					[sibling distance=0.75cm]
					node (ad) [primitive]{}{
						child[primitiveArc]{ 
							[sibling distance=4cm]
							node (adb) [primitive]{}{
								[sibling distance=0.3cm]
								child[primitiveContArc]
							}
							edge from parent node[left]{$b$}
						}
						child[primitiveContArc]
					}
					edge from parent node[left]{$d$}
				}
				child[primitiveContArc]
			}
			edge from parent node[left]{$a$}
		}
			child[primitiveArc]{ 
				[sibling distance=1cm]
				node (b) [primitive]{}{
					child[primitiveArc]{ 
						[sibling distance=4cm]
						node (ba) [primitive]{}{
							[sibling distance=0.3cm]
							child[primitiveContArc]
						}
						edge from parent node[left]{$a$}
					}
					child[primitiveArc]{ 
						[sibling distance=0.75cm]
						node (bc) [primitive]{}{
							child[primitiveArc]{ 
								[sibling distance=0.6cm]
								node (bca) [primitive]{}{
									child[primitiveArc]{ 
										[sibling distance=4cm]
										node (bdac) [primitive]{}{
											[sibling distance=0.3cm]
											child[primitiveContArc]
										}
										edge from parent node[left]{$c$}
									}
									child[primitiveArc]{ 
										[sibling distance=4cm]
										node (bdae) [primitive]{}{
											child[primitiveArc]{ 
												[sibling distance=4cm]
												node (bdaec) [primitive]{}{
												}
												edge from parent node[left]{$c$}
											}
										}
										edge from parent node[left]{$e$}
									}
									child[primitiveContArc]
								}
								edge from parent node[left]{$a$}
							}
							child[primitiveArc]{ 
								[sibling distance=4cm]
								node (bcd) [primitive]{}{
									[sibling distance=0.3cm]
									child[primitiveContArc]
								}
								edge from parent node[left]{$c$}
							}
							child[primitiveContArc]
						}
						edge from parent node[left]{$c$}
					}
					child[primitiveArc]{ 
						[sibling distance=4cm]
						node (abd) [primitive]{}{
							child[primitiveContArc]
						}
						edge from parent node[left]{$d$}
					}
					child[primitiveArc]{ 
						[sibling distance=4cm]
						node (abe) [primitive]{}{
							child[primitiveArc]{ 
								[sibling distance=4cm]
								node (bea) [primitive]{}{
									[sibling distance=0.3cm]
									child[primitiveContArc]
								}
								edge from parent node[left]{$d$}
							}
						}
						edge from parent node[left]{$e$}
					}
				}
				edge from parent node[left]{$b$}
			}
			child[primitiveArc]{ 
				[sibling distance=1cm]
				node (ac) [primitive]{}{
					child[primitiveArc]{ 
						[sibling distance=4cm]
						node (ca) [primitive]{}{
							[sibling distance=0.3cm]
							child[primitiveContArc]
						}
						edge from parent node[left]{$a$}
					}
					child[primitiveArc]{ 
						[sibling distance=4cm]
						node (acb) [primitive]{}{
							[sibling distance=0.3cm]
							child[primitiveContArc]
						}
						edge from parent node[left]{$b$}
					}
					child[primitiveContArc]
				}
				edge from parent node[left]{$c$}
			}
			child[primitiveArc]{ 
				[sibling distance=1cm]
				node (ad) [primitive]{}{
					child[primitiveArc]{ 
						[sibling distance=4cm]
						node (da) [primitive]{}{
							[sibling distance=0.3cm]
							child[primitiveContArc]
						}
						edge from parent node[left]{$a$}
					}
					child[primitiveArc]{ 
						[sibling distance=4cm]
						node (adb) [primitive]{}{
							[sibling distance=0.3cm]
							child[primitiveContArc]
						}
						edge from parent node[left]{$b$}
					}
					child[primitiveArc]{ 
						[sibling distance=0.75cm]
						node (adc) [primitive]{}{
							child[primitiveArc]{ 
								[sibling distance=4cm]
								node (bca) [primitive]{}{
									[sibling distance=0.3cm]
									child[primitiveContArc]
								}
								edge from parent node[left]{$a$}
							}
							child[primitiveArc]{ 
								[sibling distance=4cm]
								node (bcd) [primitive]{}{
									[sibling distance=0.3cm]
									child[primitiveContArc]
								}
								edge from parent node[left]{$b$}
							}
						}
						edge from parent node[left]{$c$}
					}
					child[primitiveContArc]
				}
				edge from parent node[left]{$d$}
			}
			child[primitiveArc]{ 
				[sibling distance=1cm]
				node (ae) [primitive]{}{
					child[primitiveArc]{ 
						[sibling distance=4cm]
						node (aeb) [primitive]{}{
							[sibling distance=0.3cm]
							child[primitiveContArc]
						}
						edge from parent node[left]{$a$}
					}
					child[primitiveContArc]
				}
				edge from parent node[left]{$e$}
			}
};
\end{tikzpicture}  \label{fig:t0} 
	\captionsetup{width=.9\linewidth}
	\caption{MCTS for five jobs $a,b,c,d$ and $e$. The dashed lines represent random roll-outs to estimate the values of a node.}%
	\label{fig:tree}%
\end{figure}
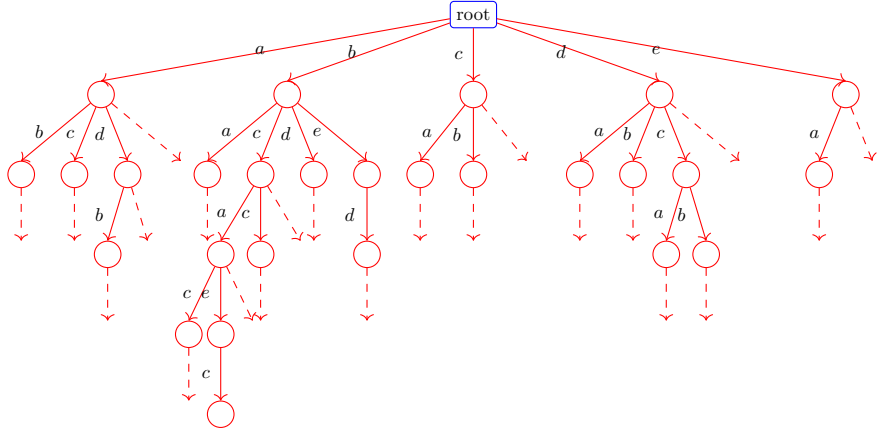

When the selection strategy chooses a state, which has not been stored in the trees, a \textit{roll-out} is performed from there onwards. Not-yet-planned jobs are randomly ordered and sequentially appended to the current partial plan, forming a complete plan at the moment the roll-out reaches a terminal state. Next, the tree is \textit{expanded} by adding as a node the first state encountered in the roll-out. Subsequently, the completed plan is evaluated. The value of this plan is \textit{backpropagated} through the previously traversed nodes, updating the average score and number of visits of all ancestor nodes. 
An example of a MCTS tree with five jobs is shown in Figure \ref{fig:tree}. 

In practice, due to limited computation time, instead of expanding the tree fully until the terminal states, the search is given a time limit $T$ to find the best child at the current root node. Once the given time is exhausted, this child is chosen and its associated job is planned.
In the next step, the search is restarted at this child node. In this way, the search can focus earlier on promising parts of the tree. This method is similar to the turn-based single-player MCTS proposed by \cite{schadd2012single}.

\subsection{MCTS Enhancements}\label{subsec:enhancements}
The following four enhancements have been added to  MCTS in order  improve search performance.

%\subsubsection{Tree Reuse}
%After the search at each partial plan, the branch at the planned job explored before is reused. 
%This is similar to tree reuse proposed by \cite{pepels2014real}.

\subsubsection{Bi-gram Progressive History}
With progressive history \cite{nijssen2010enhancements}, a history is maintained regarding past actions and the scores that they lead to during roll-outs.
The assumption is that some actions are generally better regardless of their positions in the search tree.
To incorporate progressive history, the original UCT formula in Equation \ref{eq:uct} is extended to
\begin{equation}
	v_i=s_i + C  \sqrt{\frac{\ln(n_p)}{n_i}} + s_a  \frac{W}{n_i-s_i+1},
	\label{eq:progressive}
\end{equation}
where $s_a$ is the score of action $a$ in the history and $W$ is a scaling constant controlling the impact of the progressive history on node selection. This positively biases the selection of actions that previously led to higher scores.

For the planning problem, however, the original formulation of progressive history is not applicable, since each available action must be taken once to form a complete plan.
The notion of a generally better action is therefore less relevant. An adaptation based on action pairs is made to still incorporate past knowledge.
Progressive history is kept regarding the values of consecutive action pairs of actions, called bi-grams instead of single actions.
Using bi-grams requires a minor adjustment in Equation \ref{eq:progressive} by using the history value of the pair of previous and current action rather than the current action alone. Note that the usage of bi-grams is a special case of N-grams by \cite{tak2012n}.

\subsubsection{Best Path Memorisation}
As the environment is deterministic, once a leaf node is reached and evaluated, 
its values is known for certain. The certainty outweighs any simulation result. Similar to \cite{cazenave2009nested}, this fact is utilised in best path memorisation. In case of reaching a terminal state during roll-out, which evaluates higher than all previously observed terminal states, the value and the sequence of actions leading the terminal node is memorised. When the final plan has to be chosen in the end, instead of choosing nodes with the highest average child, the nodes from the memorised path are chosen.

\subsubsection{Time Redistribution}\label{subsec:adjtime}
As the search towards terminal nodes progressively creates the full plan, the number of to-be-planned jobs reduces as the partial plans contains more jobs.
Given the total allowable computation time, each search step is allocated with different amount of time based on the number of jobs yet to be planned. 
The redistribution for time $t(d)$ of step $d$ is given in Equation \ref{distribution} as follows.

\begin{equation}
	t(d) = \frac{-1.8T}{|J|} \times d + 1.9T
	\label{distribution}
\end{equation}

\noindent where parameter $T$ is set 10 seconds and $|J|$ represents the total number of job.

\subsubsection{Parallelisation}
Robust performance of MCTS relies on large numbers of random simulations.
It is therefore important to achieve as many simulations as possible within the given computation time.
Parallelisation is common in MCTS \cite{segal2010scalability}. Two ways of parallelisation are used, called \textit{root parallelisation} and \textit{tree parallelisation}.

In root parallelisation \cite{chaslot2008parallel}, each thread is given a different random seed and grows a separate MCTS tree.
The job leading to the plan with the highest value among all MCTS trees is then chosen to be planed next. 

%Here, not all threads start with the same tree. In case the best child deviated from the best path, one thread follows the best child instead of the best path. Another thread does not perform tree reuse but starts with an empty tree. All remaining threads create copies of the best child's branch and perform MCTS with tree reuse as introduced above. Due to no communication between the threads during search, root parallelisation can be used on distributed memory systems. %%best child's branch?  Do you mean the best path

In tree parallelisation \cite{chaslot2008parallel}, all  threads operate on the same tree. 
They each perform selection, roll-out and expansion simultaneously. In tree parallelisation, during the backpropagation of roll-out results, multiple threads might access the same node. To avoid this, node updates in the backpropagation are locked such that one thread at the time can access the node's data. Due to the sharing of information between node, tree parallelisation is applicable on shared memory systems. As the evaluation is slow, such parallelisation utilises CPU better and leads a larger lookahead of the search. 

\subsection{Hierarchical Monte-Carlo Tree Search}\label{subsec:HMCTS}

While the search enhancements partially circumvent problem caused by
the large number of jobs, they do not reduce the branching factor. 
Hierarchical Monte-Carlo Tree Search (H-MCTS) circumvents the problem of large branching factor by hierarchical planning with job abstraction.
The search starts at the most abstract job, which is the root of the hierarchy. 
At each node, a nested MCTS is run on more concrete options under the hierarchy.
This is executed recursively up till the primitive jobs.

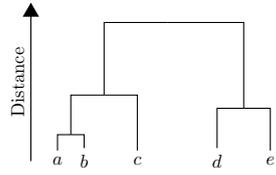
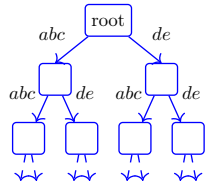
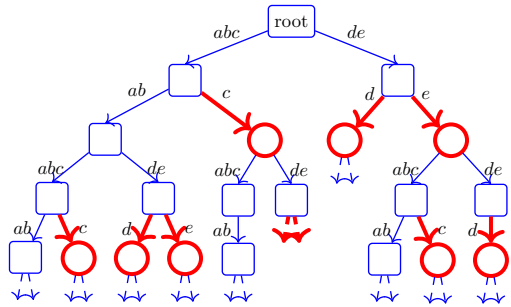
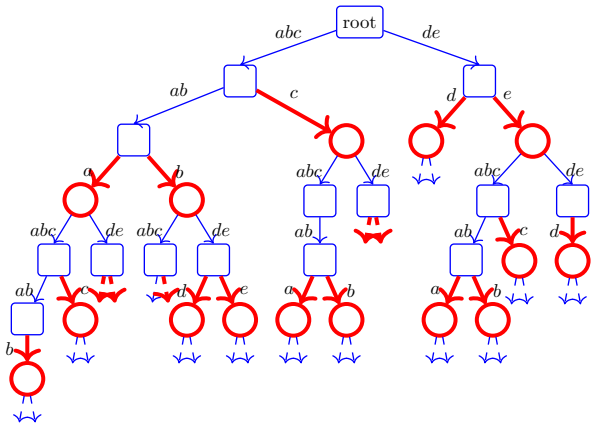
\begin{figure}%
	\centering
	%\qquad
	\subfloat[Hierarchical option abstraction.]{\begin{tikzpicture}[sloped, scale=0.7, every node/.style={scale=0.7}]
\node (a) at (2,0) {$a$};
\node (b) at (2.5,0) {$b$};
\node (c) at (3.5,0) {$c$};
\node (d) at (5,0) {$d$};
\node (e) at (6,0) {$e$};
\node (ab) at (2.25,0.5) {};
\node (abc) at (2.875,1.25) {};

\node (de) at (5.5,1) {};
\node (all) at (4.0625,2.625) {};

\draw  (a) |- (ab.center);
\draw  (b) |- (ab.center);
\draw  (c) |- (abc.center);
\draw  (ab.center) |- (abc.center);

\draw  (d) |- (de.center);
\draw  (e) |- (de.center);

\draw  (abc.center) |- (all.center);
\draw  (de.center) |- (all.center);

\draw[->,-triangle 60] (1.5,0) -- node[above]{Distance} (1.5,3);
\end{tikzpicture} \label{fig:ht}} \\
	\subfloat[Search tree on options $abc$ and $de$.]{\begin{tikzpicture}[
scale=0.7, every node/.style={scale=0.7},
abstract/.style={rectangle, draw=blue, rounded corners=0.5mm, fill=none,line width=0.5pt,
	text centered, anchor=north, text=black, minimum size=6mm},
abstractArc/.style={rectangle, draw=blue, rounded corners=0.5mm, fill=none,line width=0.5pt,
	text centered, anchor=north, text=black, minimum size=6mm},
root/.style={circle, draw=blue, fill=none,line width=0.5pt,
	text centered, anchor=north, text=black},
primitive/.style={circle, draw=red, fill=none,line width=1.5pt,
	text centered, anchor=north, text=black, minimum size=6mm},
primitiveArc/.style={circle, draw=red, fill=none,line width=1.5pt,
	text centered, anchor=north, text=black, minimum size=6mm},
contArc/.style={dashed, draw=black, fill=none,
	text centered, anchor=north, text=black, minimum size=6mm},
abstractContArc/.style={dashed, draw=blue, fill=none,line width=0.5pt,
	text centered, anchor=north, text=black, minimum size=6mm},
primitiveContArc/.style={dashed, draw=red, fill=none,line width=1.5pt,
	text centered, anchor=north, text=black, minimum size=6mm},
level distance=0.5cm, growth parent anchor=south
	]
	\node (root) [abstract] {root} [->]{ 
		[sibling distance=2cm]
		child[abstractArc]{ 
			[sibling distance=1cm]
			node (abc0) [abstract]{}{
				child[abstractArc]{ 
					node (abc1) [abstract]{}{
						[sibling distance=0.3cm]
						child[abstractContArc]
						child[abstractContArc]
					}
					edge from parent node[above left]{$abc$}
				}
				child[abstractArc]{ 
				node (de1) [abstract]{}{
					[sibling distance=0.3cm]
					child[abstractContArc]
					child[abstractContArc]
				}
				edge from parent node[above right]{$de$}
			}
			}
			edge from parent node[above left]{$abc$}
		}
		child[abstractArc]{ 
			[sibling distance=1cm]
			node (de0) [abstract]{}{
				child[abstractArc]{ 
					node (abc2) [abstract]{}{
						[sibling distance=0.3cm]
						child[abstractContArc]
						child[abstractContArc]
					}
					edge from parent node[above left]{$abc$}
				}
				child[abstractArc]{ 
					node (de2) [abstract]{}{
						[sibling distance=0.3cm]
						child[abstractContArc]
						child[abstractContArc]
					}
					edge from parent node[ above right]{$de$}
				}
			}
			edge from parent node[above right]{$de$}
		}
	};

	\end{tikzpicture} \label{fig:tb}} \\
	\qquad
	\subfloat[First nesting level on options $abc(ab,c)$ and $de(d,e)$.]{\begin{tikzpicture}[
scale=0.7, every node/.style={scale=0.7},
abstract/.style={rectangle, draw=blue, rounded corners=0.5mm, fill=none,line width=0.5pt,
	text centered, anchor=north, text=black, minimum size=6mm},
abstractArc/.style={rectangle, draw=blue, rounded corners=0.5mm, fill=none,line width=0.5pt,
	text centered, anchor=north, text=black, minimum size=6mm},
root/.style={circle, draw=blue, fill=none,line width=0.5pt,
	text centered, anchor=north, text=black},
primitive/.style={circle, draw=red, fill=none,line width=1.5pt,
	text centered, anchor=north, text=black, minimum size=6mm},
primitiveArc/.style={circle, draw=red, fill=none,line width=1.5pt,
	text centered, anchor=north, text=black, minimum size=6mm},
contArc/.style={dashed, draw=black, fill=none,
	text centered, anchor=north, text=black, minimum size=6mm},
abstractContArc/.style={dashed, draw=blue, fill=none,line width=0.5pt,
	text centered, anchor=north, text=black, minimum size=6mm},
primitiveContArc/.style={dashed, draw=red, fill=none,line width=1.5pt,
	text centered, anchor=north, text=black, minimum size=6mm},
level distance=0.5cm, growth parent anchor=south
]
\node (root) [abstract] {root} [->]{ 
	[sibling distance=4cm]
	child[abstractArc]{ 
		[sibling distance=3cm]
		node (abc0) [abstract]{}{
			child[abstractArc]{ 
				[sibling distance=2cm]
				node (ab0) [abstract]{}{
					child[abstractArc]{ 
							[sibling distance=1cm]
								[sibling distance=1cm]
								node (abc1) [abstract]{}{
									child[abstractArc]{ 
										node (ab1) [abstract]{}{
											[sibling distance=0.3cm]
											child[abstractContArc]
											child[abstractContArc]
										}
										edge from parent node[ left]{$ab$}
									}
									child[primitiveArc]{ 
										node (c1) [primitive]{}{
											[sibling distance=0.3cm]
											child[abstractContArc]
											child[abstractContArc]
										}
										edge from parent node[ right]{$c$}
									}
								}
								edge from parent node[left]{$abc$}
							}
							child[abstractArc]{ 
								[sibling distance=1cm]
								node (de1) [abstract]{}{
									child[primitiveArc]{ 
										node (d2) [primitive]{}{
											[sibling distance=0.3cm]
											child[abstractContArc]
											child[abstractContArc]
										}
										edge from parent node[ left]{$d$}
									}
									child[primitiveArc]{ 
										node (e2) [primitive]{}{
											[sibling distance=0.3cm]
											child[abstractContArc]
											child[abstractContArc]
										}
										edge from parent node[ right]{$e$}
									}
								}
								edge from parent node[ right]{$de$}
							}
				}
				edge from parent node[above]{$ab$}
			}
			child[primitiveArc]{ 
				node (c0) [primitive]{}{
					[sibling distance=1cm]
					child[abstractArc]{ 
						[sibling distance=1cm]
						node (abc3) [abstract]{}{
							child[abstractArc]{ 
								node (ab3) [abstract]{}{
									[sibling distance=0.3cm]
									child[abstractContArc]
									child[abstractContArc]
								}
								edge from parent node[left]{$ab$}
							}
						}
						edge from parent node[ left]{$abc$}
					}
					child[abstractArc]{ 
						[sibling distance=1cm]
						node (de3) [abstract]{}{
							[sibling distance=0.3cm]
							child[primitiveContArc]
							child[primitiveContArc]
							\iffalse
							child[primitiveArc]{ 
								node (d3) [primitive]{}{
								}
								edge from parent node[ left]{$d$}
							}
							child[primitiveArc]{ 
								node (e3) [primitive]{}{
								}
								edge from parent node[ right]{$e$}
							}
							\fi
						}
						edge from parent node[ right]{$de$}
					}
				}
				edge from parent node[ above]{$c$}
			}
		}
		edge from parent node[above]{$abc$}
	}
	child[abstractArc]{ 
		[sibling distance=2cm]
		node (de0) [abstract]{}{
			child[primitiveArc]{ 
				node (d0) [primitive]{}{
					[sibling distance=0.3cm]
					child[abstractContArc]
					child[abstractContArc]
				}
				edge from parent node[ above]{$d$}
			}
			child[primitiveArc]{ 
				node (e0) [primitive]{}{
					[sibling distance=1.5cm]
					child[abstractArc]{ 
						[sibling distance=1cm]
						node (abc5) [abstract]{}{
							child[abstractArc]{ 
								node (ab5) [abstract]{}{
									[sibling distance=0.3cm]
									child[abstractContArc]
									child[abstractContArc]
								}
								edge from parent node[ left]{$ab$}
							}
							child[primitiveArc]{ 
								node (c5) [primitive]{}{
									[sibling distance=0.3cm]
									child[abstractContArc]
									child[abstractContArc]
								}
								edge from parent node[ right]{$c$}
							}
						}
						edge from parent node[ left]{$abc$}
					}
					child[abstractArc]{ 
						[sibling distance=1cm]
						node (de5) [abstract]{}{
							child[primitiveArc]{ 
								node (d5) [primitive]{}{
									[sibling distance=0.3cm]
									child[abstractContArc]
									child[abstractContArc]
								}
								edge from parent node[left]{$d$}
							}
						}
						edge from parent node[ right]{$de$}
					}
				}
				edge from parent node[ above]{$e$}
			}
		}
		edge from parent node[above]{$de$}
	}
};
\end{tikzpicture}  \label{fig:t1}} \\
	\subfloat[Second nesting level on options $abc(ab(a,b),c)$ and \newline $de(d,e)$.]{\begin{tikzpicture}[
scale=0.7, every node/.style={scale=0.7},
abstract/.style={rectangle, draw=blue, rounded corners=0.5mm, fill=none,line width=0.5pt,
	text centered, anchor=north, text=black, minimum size=6mm},
abstractArc/.style={rectangle, draw=blue, rounded corners=0.5mm, fill=none,line width=0.5pt,
	text centered, anchor=north, text=black, minimum size=6mm},
root/.style={circle, draw=blue, fill=none,line width=0.5pt,
	text centered, anchor=north, text=black},
primitive/.style={circle, draw=red, fill=none,line width=1.5pt,
	text centered, anchor=north, text=black, minimum size=6mm},
primitiveArc/.style={circle, draw=red, fill=none,line width=1.5pt,
	text centered, anchor=north, text=black, minimum size=6mm},
contArc/.style={dashed, draw=black, fill=none,
	text centered, anchor=north, text=black, minimum size=6mm},
abstractContArc/.style={dashed, draw=blue, fill=none,line width=0.5pt,
	text centered, anchor=north, text=black, minimum size=6mm},
primitiveContArc/.style={dashed, draw=red, fill=none,line width=1.5pt,
	text centered, anchor=north, text=black, minimum size=6mm},
level distance=0.5cm, growth parent anchor=south
]
\node (root) [abstract] {root} [->]{ 
	[sibling distance=4.5cm]
	child[abstractArc]{ 
		[sibling distance=4cm]
		node (abc0) [abstract]{}{
			child[abstractArc]{ 
				[sibling distance=2cm]
				node (ab0) [abstract]{}{
					child[primitiveArc]{ 
						node (a0) [primitive]{}{
							[sibling distance=1cm]
							child[abstractArc]{ 
								[sibling distance=1cm]
								node (abc1) [abstract]{}{
									child[abstractArc]{ 
										node (ab1) [abstract]{}{
											\iffalse
											child[primitiveArc]{ 
												node (a1) [primitive]{}{
												}
												edge from parent node[ left]{$a$}
											}
											\fi
											child[primitiveArc]{ 
												node (b1) [primitive]{}{
													[sibling distance=0.3cm]
													child[abstractContArc]
													child[abstractContArc]
												}
												edge from parent node[left]{$b$}
											}
										}
										edge from parent node[ left]{$ab$}
									}
									child[primitiveArc]{ 
										node (c1) [primitive]{}{
											[sibling distance=0.3cm]
											child[abstractContArc]
											child[abstractContArc]
										}
										edge from parent node[ right]{$c$}
									}
								}
								edge from parent node[left]{$abc$}
							}
							child[abstractArc]{ 
								[sibling distance=1cm]
								node (de1) [abstract]{}{
									[sibling distance=0.3cm]
									child[primitiveContArc]
									child[primitiveContArc]
									\iffalse
									child[primitiveArc]{ 
										node (d1) [primitive]{}{
										}
										edge from parent node[ left]{$d$}
									}
									child[primitiveArc]{ 
										node (e1) [primitive]{}{
										}
										edge from parent node[ right]{$e$}
									}
									\fi
								}
								edge from parent node[ right]{$de$}
							}
						}
						edge from parent node[ left]{$a$}
					}
					child[primitiveArc]{ 
						node (b0) [primitive]{}{
							[sibling distance=1cm]
							child[abstractArc]{ 
								[sibling distance=1cm]
								node (abc2) [abstract]{}{
									[sibling distance=0.3cm]
									child[abstractContArc]
									child[primitiveContArc]
									\iffalse
									child[abstractArc]{ 
										node (ab2) [abstract]{}{
											child[primitiveArc]{ 
												node (a2) [primitive]{}{
												}
												edge from parent node[left]{$a$}
											}
										}
										edge from parent node[ left]{$ab$}
									}
									child[primitiveArc]{ 
										node (c2) [primitive]{}{
										}
										edge from parent node[ right]{$c$}
									}
									\fi
								}
								edge from parent node[ left]{$abc$}
							}						
							child[abstractArc]{ 
								[sibling distance=1cm]
								node (de2) [abstract]{}{
									child[primitiveArc]{ 
										node (d2) [primitive]{}{
											[sibling distance=0.3cm]
											child[abstractContArc]
											child[abstractContArc]
										}
										edge from parent node[ left]{$d$}
									}
									child[primitiveArc]{ 
										node (e2) [primitive]{}{
											[sibling distance=0.3cm]
											child[abstractContArc]
											child[abstractContArc]
										}
										edge from parent node[ right]{$e$}
									}
								}
								edge from parent node[ right]{$de$}
							}
						}
						edge from parent node[ right]{$b$}
					}
				}
				edge from parent node[above]{$ab$}
			}
			child[primitiveArc]{ 
				node (c0) [primitive]{}{
					[sibling distance=1cm]
					child[abstractArc]{ 
						[sibling distance=1cm]
						node (abc3) [abstract]{}{
							child[abstractArc]{ 
								node (ab3) [abstract]{}{
									child[primitiveArc]{ 
										node (a3) [primitive]{}{
											[sibling distance=0.3cm]
											child[abstractContArc]
											child[abstractContArc]
										}
										edge from parent node[ left]{$a$}
									}
									child[primitiveArc]{ 
										node (b3) [primitive]{}{
											[sibling distance=0.3cm]
											child[abstractContArc]
											child[abstractContArc]
										}
										edge from parent node[ right]{$b$}
									}
								}
								edge from parent node[left]{$ab$}
							}
						}
						edge from parent node[ left]{$abc$}
					}
					child[abstractArc]{ 
						[sibling distance=1cm]
						node (de3) [abstract]{}{
							[sibling distance=0.3cm]
							child[primitiveContArc]
							child[primitiveContArc]
							\iffalse
							child[primitiveArc]{ 
								node (d3) [primitive]{}{
								}
								edge from parent node[ left]{$d$}
							}
							child[primitiveArc]{ 
								node (e3) [primitive]{}{
								}
								edge from parent node[ right]{$e$}
							}
							\fi
						}
						edge from parent node[ right]{$de$}
					}
				}
				edge from parent node[ above]{$c$}
			}
		}
		edge from parent node[above]{$abc$}
	}
	child[abstractArc]{ 
		[sibling distance=2cm]
		node (de0) [abstract]{}{
			child[primitiveArc]{ 
				node (d0) [primitive]{}{
					[sibling distance=0.3cm]
					child[abstractContArc]
					child[abstractContArc]
					\iffalse
					[sibling distance=1.5cm]
					child[abstractArc]{ 
						[sibling distance=1cm]
						node (abc4) [abstract]{}{
							\iffalse
							child[abstractArc]{ 
								node (ab4) [abstract]{}{
									child[primitiveArc]{ 
										node (a4) [primitive]{}{
											[sibling distance=0.3cm]
											child[abstractContArc]
											child[abstractContArc]
										}
										edge from parent node[ left]{$a$}
									}
									child[primitiveArc]{ 
										node (b4) [primitive]{}{
											[sibling distance=0.3cm]
											child[abstractContArc]
											child[abstractContArc]
										}
										edge from parent node[ right]{$b$}
									}
								}
								edge from parent node[ left]{$ab$}
							}
							child[primitiveArc]{ 
								node (c4) [primitive]{}{
									[sibling distance=0.3cm]
									child[abstractContArc]
									child[abstractContArc]
								}
								edge from parent node[ right]{$c$}
							}
							\fi
						}
						edge from parent node[ left]{$abc$}
					}
					child[abstractArc]{ 
						[sibling distance=1cm]
						node (de4) [abstract]{}{
							child[primitiveArc]{ 
								node (e4) [primitive]{}{
								}
								edge from parent node[left]{$e$}
							}
						}
						edge from parent node[ right]{$de$}
					}
					\fi
				}
				edge from parent node[ above]{$d$}
			}
			child[primitiveArc]{ 
				node (e0) [primitive]{}{
					[sibling distance=1.5cm]
					child[abstractArc]{ 
						[sibling distance=1cm]
						node (abc5) [abstract]{}{
							child[abstractArc]{ 
								node (ab5) [abstract]{}{
									child[primitiveArc]{ 
										node (a5) [primitive]{}{
											[sibling distance=0.3cm]
											child[abstractContArc]
											child[abstractContArc]
										}
										edge from parent node[ left]{$a$}
									}
									child[primitiveArc]{ 
										node (b5) [primitive]{}{
											[sibling distance=0.3cm]
											child[abstractContArc]
											child[abstractContArc]
										}
										edge from parent node[ right]{$b$}
									}
								}
								edge from parent node[ left]{$ab$}
							}
							child[primitiveArc]{ 
								node (c5) [primitive]{}{
									[sibling distance=0.3cm]
									child[abstractContArc]
									child[abstractContArc]
								}
								edge from parent node[ right]{$c$}
							}
						}
						edge from parent node[ left]{$abc$}
					}
					child[abstractArc]{ 
						[sibling distance=1cm]
						node (de5) [abstract]{}{
							child[primitiveArc]{ 
								node (d5) [primitive]{}{
									[sibling distance=0.3cm]
									child[abstractContArc]
									child[abstractContArc]
								}
								edge from parent node[left]{$d$}
							}
						}
						edge from parent node[ right]{$de$}
					}
				}
				edge from parent node[ above]{$e$}
			}
		}
		edge from parent node[above]{$de$}
	}
};
\end{tikzpicture}  \label{fig:t}} \\
	\captionsetup{width=.9\linewidth}
	\caption{Nested search over abstraction hierarchy. The hierarchy is shown in Subfgure (a), and Subfigures (b) to (d) show the beginning of a search tree over given hierarchy with increasing nesting level. Blue components of the trees represent abstract jobs and thick red ones represent primitive jobs.}%
	\label{fig:htree}%
\end{figure}

The procedure is further illustrated with an example in Figure \ref{fig:htree}, where blue indicates the execution of abstract jobs, and red that of primitive jobs.
Figure \ref{fig:ht} shows the hierarchy of the five primitive jobs $a,b,c,d$ and $e$, where two highest-level abstract jobs exists, namely $abc$ and $de$.
Figure \ref{fig:tb} shows the most abstract search level with only abstract jobs $abc$ and $de$.
Figure \ref{fig:t1} shows the first nesting level, where the option space $abc$ is searched by more concrete options $ab$ and $c$. The same goes with $de$ and its more concrete jobs $d$ and $e$. 
%TODO reformulate
Here $c,d$ and $e$ are primitive jobs and are represented by red tree parts.
%END TODO reformulate
Figure \ref{fig:t} shows the second nesting level where option space $ab$ is searched by more concrete jobs $a$ and $b$. All nested searches of abstract jobs now terminate because of reaching primitive jobs.

\subsection{Job Abstraction}\label{subsec:jobAbstraction}
Job Abstraction for H-MCTS can be achieved by grouping similar ones into the same abstract job. This subsection proposed two abstraction approaches. The first creates the abstraction via hierarchical clustering and then provides the clusters to the search as input (\ref{sec:preprocess}). Because the abstraction and search are two separated steps, this approach is referred to as \textit{detached} abstraction. The second integrates the abstraction in the search and therefore is called \textit{integrated} abstraction (\ref{sec:automated}). 

\subsubsection{Detached Abstraction}\label{sec:preprocess}
Based on the features of the jobs, pairwise distance is calculated, for example
with Cosine or Euclidean distance. The features include processing time, deadline and required resources.
The jobs represented as features form the input to the hierarchical clustering procedure.
Every new option is joined with the nearest option to form an more abstract
job. This is repeated until one aggregated option is left.

\begin{figure}[h]%
	\centering
	\subfloat[First joint $a$ and $b$.]{\begin{tikzpicture}[sloped]
\node (a) at (2,0) {$a$};
\node (b) at (3,0) {$b$};
\node (c) at (4,0) {$c$};
\node (ab) at (2.5,1) {};
\node (abc) at (2.75,1.5) {};

\draw  (a) |- (ab.center);
\draw  (b) |- (ab.center);
\draw  (c) |- (abc.center);
\draw  (ab.center) |- (abc.center);

\draw[->,-triangle 60] (1,0) -- node[above]{\scriptsize Distance} (1,2);
\end{tikzpicture} \label{fig:h1}}%
	\subfloat[First joint $b$ and $c$.]{\begin{tikzpicture}[sloped]
\node (a) at (2,0) {$a$};
\node (b) at (3,0) {$b$};
\node (c) at (4,0) {$c$};
\node (bc) at (3.5,1) {};
\node (abc) at (2.75,1.5) {};

\draw  (c) |- (bc.center);
\draw  (b) |- (bc.center);
\draw  (a) |- (abc.center);
\draw  (bc.center) |- (abc.center);

\draw[->,-triangle 60] (1,0) -- node[above]{\scriptsize Distance} (1,2);
\end{tikzpicture}  \label{fig:h2}}%
	\captionsetup{width=.9\linewidth}
	\caption{Hierarchical cluster of three options $a,b$ and $c$ where dist($a$,$b$) = dist($b$,$c$) and dist($a$,$c$) = $2$ dist($a$,$b$).}%
	\label{fig:hierarchy}%
\end{figure}
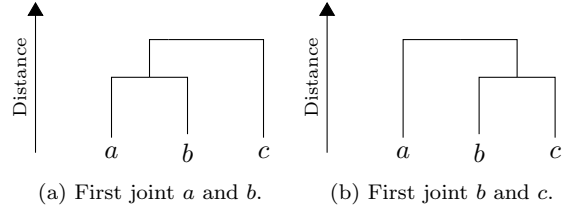

Note, however, that hierarchical clustering is not stable with regard to input orders. % left to right
Figures \ref{fig:h1} and \ref{fig:h2} show an example of hierarchical clusters of three options $a$, $b$ and $c$, where dist($a$,$b$) = dist($b$,$c$) and dist($a$,$c$) = $2 \cdot$ dist($a$,$b$). 
When the input order is $[a,b,c]$, the hierarchy in Figure \ref{fig:h1} is the output. But for the case when the input order becomes $[c,b,a]$, the results is depicted in Figure \ref{fig:h2}. 
Both are valid abstractions, but might lead to different results during planning.

\subsubsection{ Integrated Abstraction}\label{sec:automated}
Preprocessed abstraction via hierarchical clustering has the downside that the output is not stable with regard to input ordering, as exemplified previously.
Since the constructed abstraction has a large impact on the search outcome, a more robust approach is needed.

Firstly, initial clusters are created with mean-shift clustering \cite{comaniciu2002mean} on all jobs to initially reduce complexity. 
%TODO reformulate
%[\textbf{why mean-shift: can find poles of distributions, no assumption on number of clusters.}]
This generates an initial abstraction level ensuring that equal or  similar jobs are considered in the same option. Mean-shift does not make assumptions on number of clusters contrary to other methods such as $k$-mean clustering.
%END TODO reformulate
Instead of joining the initial clusters based on distance, the hierarchy formation is incorporated and automatically handled integrated in the H-MCTS.
The search then consists of two parts; an abstraction search is added to the existing planning search.
In the abstraction search, H-MCTS searches which options to hierarchically join until one abstract job is left.
For example, consider the previously used illustration in Figure \ref{fig:hierarchy}, where all three possible hierarchies are shown. 
To arrive at the result of Figure \ref{fig:h1}, H-MCTS starts with three options $a,b$ and $c$. 
Abstracting $a$ and $b$ into $(a,b)$ is evaluated better by running Monte-Carlo simulations than $(b,c)$. Therefore, the best child in the first ply is $(a,b)$.
In the second ply, the only child is abstracting $(a,b)$ and $c$ to $((a,b),c)$.
%TODO reformulate
This brings to an abstraction result of having most abstract job $abc$, with more concrete options $ab$ and $c$, where $ab$ contains of the primitive jobs $a$ and $b$ as illustrated in Figure \ref{fig:h1}.
%END TODO reformulate
After this abstraction search, the planning search operates in the same way as introduced before.

Noteworthy is that this approach is modular. Whether the used search framework is MCTS or NMCS, the abstraction module does not need to change.

\subsection{Modelling the Scheduling Heuristic}\label{sec:learning}
The output from the scheduling heuristic in terms of completion times of the planned jobs is required by the MCTS as an evaluation metric for the performed simulations. As MCTS rests upon large numbers of random samples, these evaluations must be fast.

\begin{figure}%
	\centering
	%\subfloat[Convolutional autoencoder with LSTM latent space: Input active jobs at time $t$ and newly planned job at time $t$, output active jobs at time $t+1$.]{\includegraphics[width=0.4\textwidth]{Chapter3/figures/recurrentEnvSim} \label{fig:envSim}}%
	%\qquad
	%\subfloat[LSTM Language Model: Input active jobs, output softmax over all jobs giving probability of which job is completed next.]{\includegraphics[width=0.4\textwidth]{Chapter3/figures/lstm}  \label{fig:lstm}}\\%
	\includegraphics[width=\textwidth]{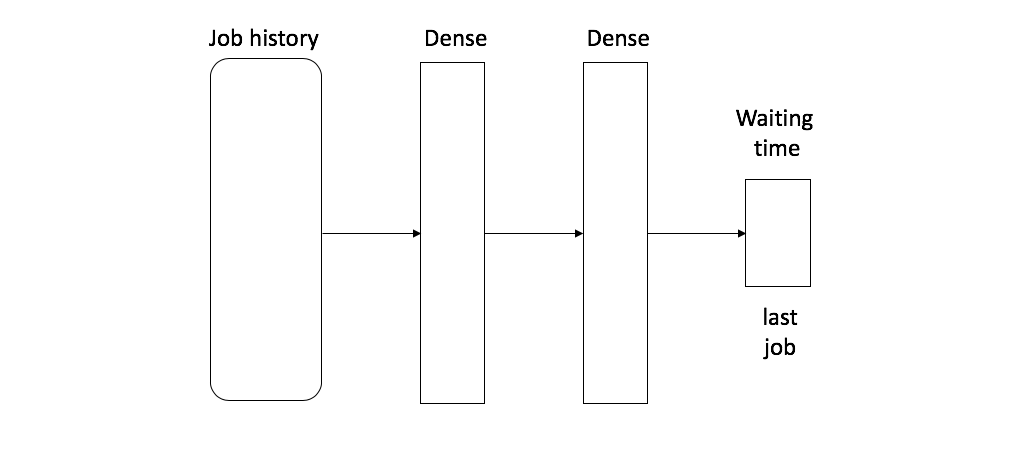}  
	
	\caption{Feed-forward neural-network architecture: input is the recent history of jobs; output is the waiting time of last added job.}%
	\label{fig:architectures}%
\end{figure}

The first issue is that the scheduling heuristic is often too time consuming to act as such an evaluation metric for Monte-Carlo simulations. The second issue is that the underlying  scheduling heuristic is encapsulated in the production controller.  As the exact mechanism of the heuristic is unknown to other parts of the system, it cannot be explicitly modelled. Therefore, the behaviour of the heuristic scheduler has to be learned. This could be realised by training a neural network, which would be relatively fast as it is simply a forward pass through the network. As the production processes are sequential, it is reasonable to opt for recurrent neural network (RNN), which has shown strong performance in modelling sequences \cite{chung2014empirical}.  Examples architectures  are LSTMs, which are particularly strong in modelling long-term dependencies, or in other words, remembering past states of the sequence. However, the long-term ``memory'' may not be required in this case, because completed jobs do not influence the line anymore. The memory span of the network only needs to cover as many jobs as can be processed in parallel. In this light, a less sophisticated architecture is used. A small feed forward network, as shown in Figure \ref{fig:architectures}, takes the most recently planned jobs as input, and seeks to predict the waiting time of the last added job.

Experiments revealed that feed-forward architecture has an accuracy of 95.3\%  and 94.2\% during training and testing, respectively, compared to a  LSTM-based architecture with an accuracy of 97.9\% and 96.2\% \cite{wimmenauer2019}. The fact that such a simple network structure still achieves reasonable accuracy confirms the hypothesis that the long-term-dependency-modelling capabilities of LSTMs are not necessary in this case. Although the accuracy is slightly lower than the  LSTM architecture, inference is 2.4 times faster with this feed forward architecture. Therefore, this architecture is used in all upcoming experiments involving heuristic simulators.

%In this article three neural network architectures are implemented to model the behaviour of the underlying scheduling heuristic. They are as follows. 
%
%
%
%\subsubsection{Approach 1: Convolutional Autoencoder with LSTM Latent Space }
%
%The first approach uses a convolutional autoencoder with recurrent latent space, following the work of \cite{chiappa2017recurrent}. 
%The authors demonstrated the model's strong performance in predicting hundreds of future frames in Atari games.
%This structure is adapted for this work and is shown in Figure \ref{fig:envSim}. 
%The network uses active jobs at time $t$ to predict active jobs at the next time step $t+1$. 
%The active jobs are represented by their processes, and the processes are in turn encoded by one-hot vectors.
%Therefore, the input are zero-padded matrices.
%After the convolutional encoder, the input is passed into a long short-term memory (LSTM) layer. The action is appended in this layer, after which the final convolutional decoder attempts to predict the jobs at the next time step. 
%The action in this case is the job added to the plan at time step $t$.
%
%\subsubsection{Approach 2: LSTM Language Model}
%The second approach is inspired by natural language processing systems, which typically use recurrent neural networks followed by dense and softmax layers over all possible tokens to predict the next most likely token, such as in \cite{graves2005framewise}.
%The structure used for this work is shown in Figure \ref{fig:lstm}. 
%The inputs are the active jobs, represented in the same way as in Approach 1. 
%The output is a softmax over all jobs identifying the active job which will be completed next.
%
%
%\subsubsection{Approach 3: Feed Forward Neural Network }

\subsection{Modified Evaluation Metric}\label{subsec:TSE}
%% This is not quite clear. Rewrite for journal submission
When the search procedure reaches a terminal state, the resulting plan needs to be evaluated.
The evaluation can be made from multiple perspectives with regard to the completion time of all jobs $j\in J$.
The makespan $c_{\max}$ is defined $\max(c_j), \forall j \in J$.
The lateness is $L_j=c_j-d_j$, with maximum lateness being $L_{\max} = \max(L_j), \forall j \in J$, and total lateness being  $L = \sum{L_j}, \forall j \in J$.
Note that scheduling heuristics are typically designed for one particular objective, in case of the presence of a different objective, the heuristic would need to be modified or replaced. However, in the MCTS framework, by changing the evaluation metric, the scheduling heuristic can be kept unchanged in case of objective change. It is hypothesized that scheduling heuristics can exhibit behaviour that they are not designed for if the MCTS evaluation metric is modified correctly.

%%%%%%%%%%%%%%%%%%%%%%%%%%%%%%%%%%%%%%%%%%%%%%%%%%%%%%%%%%%%%%%%%%%%%%%%%%%%%%%%%
%%%   Experimental setup
%%%%%%%%%%%%%%%%%%%%%%%%%%%%%%%%%%%%%%%%%%%%%%%%%%%%%%%%%%%%%%%%%%%%%%%%%%%%%%%%%

\section{Experimental Setup}
\label{sec:experimental setup}

In this section we present the experimental setup to evaluate our approach. Subsections \ref{sec:testedJSP}, \ref{sec:benchmark}, and \ref{sec:metric} discuss the tested job shop problems, the benchmark, and reported metrics, respectively.

\subsection{Tested Job Shop Problems} 
\label{sec:testedJSP}

The proposed approach is evaluated on three types of JSPs.
The first is a synthetic JSP where the optimum schedule is known.
The second type is a class of well-studied and publicly available planning scenarios of Demirkol et al.\,\cite{demirkol1998benchmarks}. 
Third, a real-world JSP involving multiple additional constraints is also tested.
Each of the types uses a different underlying blackbox heuristic. 

\subsubsection{Problem 1: Synthetic Problem} \label{subsec:prob1}

The synthetic problem is designed to be easily understandable yet still challenging for the search to find the optimum.
More specifically, it has a straightforward suboptimal solution, which can be considered as a local minimum where the search could be entrapped.

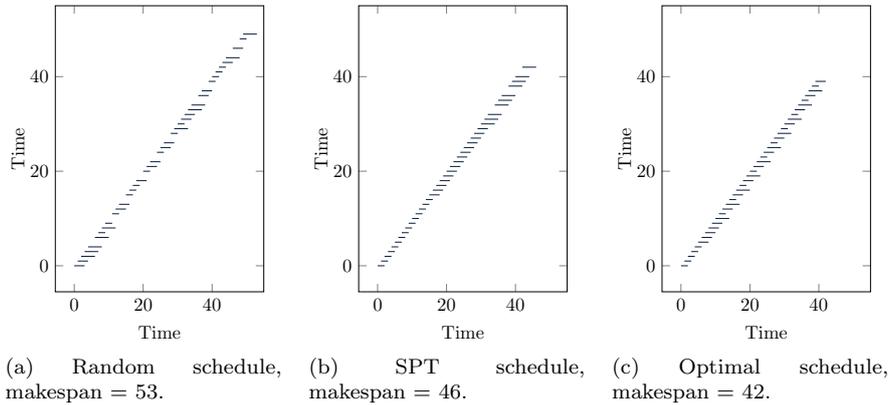
\begin{figure}%
	\centering
	\subfloat[Random schedule, makespan = 53.]{\begin{tikzpicture}[scale = 0.7]
\pgfplotsset{every axis y label/.style={
		at={(0,0.5)},
		xshift=-20pt,
		rotate=90}}
\begin{axis}[xlabel = Time, ylabel = Time, width = 5.5 cm, height = 7 cm,ymax=55,xmax=55]
\addplot[color=umBlue,mark=none] coordinates
{(0,0.0) (3,0.0)};
\addplot[color=umBlue,mark=none] coordinates
{(1,1.0) (4,1.0)};
\addplot[color=umBlue,mark=none] coordinates
{(2,2.0) (6,2.0)};
\addplot[color=umBlue,mark=none] coordinates
{(3,3.0) (7,3.0)};
\addplot[color=umBlue,mark=none] coordinates
{(4,4.0) (8,4.0)};
\addplot[color=umBlue,mark=none] coordinates
{(6,6.0) (10,6.0)};
\addplot[color=umBlue,mark=none] coordinates
{(7,7.0) (9,7.0)};
\addplot[color=umBlue,mark=none] coordinates
{(8,8.0) (12,8.0)};
\addplot[color=umBlue,mark=none] coordinates
{(9,9.0) (11,9.0)};
\addplot[color=umBlue,mark=none] coordinates
{(11,11.0) (13,11.0)};
\addplot[color=umBlue,mark=none] coordinates
{(12,12.0) (15,12.0)};
\addplot[color=umBlue,mark=none] coordinates
{(13,13.0) (16,13.0)};
\addplot[color=umBlue,mark=none] coordinates
{(15,15.0) (17,15.0)};
\addplot[color=umBlue,mark=none] coordinates
{(16,16.0) (18,16.0)};
\addplot[color=umBlue,mark=none] coordinates
{(17,17.0) (19,17.0)};
\addplot[color=umBlue,mark=none] coordinates
{(18,18.0) (21,18.0)};
\addplot[color=umBlue,mark=none] coordinates
{(20,20.0) (22,20.0)};
\addplot[color=umBlue,mark=none] coordinates
{(21,21.0) (24,21.0)};
\addplot[color=umBlue,mark=none] coordinates
{(22,22.0) (25,22.0)};
\addplot[color=umBlue,mark=none] coordinates
{(24,24.0) (26,24.0)};
\addplot[color=umBlue,mark=none] coordinates
{(25,25.0) (28,25.0)};
\addplot[color=umBlue,mark=none] coordinates
{(26,26.0) (29,26.0)};
\addplot[color=umBlue,mark=none] coordinates
{(28,28.0) (30,28.0)};
\addplot[color=umBlue,mark=none] coordinates
{(29,29.0) (33,29.0)};
\addplot[color=umBlue,mark=none] coordinates
{(30,30.0) (32,30.0)};
\addplot[color=umBlue,mark=none] coordinates
{(31,31.0) (34,31.0)};
\addplot[color=umBlue,mark=none] coordinates
{(32,32.0) (35,32.0)};
\addplot[color=umBlue,mark=none] coordinates
{(33,33.0) (37,33.0)};
\addplot[color=umBlue,mark=none] coordinates
{(34,34.0) (38,34.0)};
\addplot[color=umBlue,mark=none] coordinates
{(36,36.0) (39,36.0)};
\addplot[color=umBlue,mark=none] coordinates
{(37,37.0) (40,37.0)};
\addplot[color=umBlue,mark=none] coordinates
{(39,39.0) (41,39.0)};
\addplot[color=umBlue,mark=none] coordinates
{(40,40.0) (42,40.0)};
\addplot[color=umBlue,mark=none] coordinates
{(41,41.0) (43,41.0)};
\addplot[color=umBlue,mark=none] coordinates
{(42,42.0) (44,42.0)};
\addplot[color=umBlue,mark=none] coordinates
{(43,43.0) (46,43.0)};
\addplot[color=umBlue,mark=none] coordinates
{(44,44.0) (48,44.0)};
\addplot[color=umBlue,mark=none] coordinates
{(46,46.0) (49,46.0)};
\addplot[color=umBlue,mark=none] coordinates
{(48,48.0) (50,48.0)};
\addplot[color=umBlue,mark=none] coordinates
{(49,49.0) (53,49.0)};
\end{axis}
\end{tikzpicture} \label{fig:p1_1}}%
	\quad
	\subfloat[SPT schedule, makespan = 46.]{\begin{tikzpicture}[scale = 0.7]
\pgfplotsset{every axis y label/.style={
		at={(0,0.5)},
		xshift=-20pt,
		rotate=90}}
\begin{axis}[xlabel = Time, ylabel = Time, width = 5.5 cm, height = 7 cm,ymax=55,xmax=55]
\addplot[color=umBlue,mark=none] coordinates
{(0,0.0) (2,0.0)};
\addplot[color=umBlue,mark=none] coordinates
{(1,1.0) (3,1.0)};
\addplot[color=umBlue,mark=none] coordinates
{(2,2.0) (4,2.0)};
\addplot[color=umBlue,mark=none] coordinates
{(3,3.0) (5,3.0)};
\addplot[color=umBlue,mark=none] coordinates
{(4,4.0) (6,4.0)};
\addplot[color=umBlue,mark=none] coordinates
{(5,5.0) (7,5.0)};
\addplot[color=umBlue,mark=none] coordinates
{(6,6.0) (8,6.0)};
\addplot[color=umBlue,mark=none] coordinates
{(7,7.0) (9,7.0)};
\addplot[color=umBlue,mark=none] coordinates
{(8,8.0) (10,8.0)};
\addplot[color=umBlue,mark=none] coordinates
{(9,9.0) (11,9.0)};
\addplot[color=umBlue,mark=none] coordinates
{(10,10.0) (12,10.0)};
\addplot[color=umBlue,mark=none] coordinates
{(11,11.0) (13,11.0)};
\addplot[color=umBlue,mark=none] coordinates
{(12,12.0) (14,12.0)};
\addplot[color=umBlue,mark=none] coordinates
{(13,13.0) (15,13.0)};
\addplot[color=umBlue,mark=none] coordinates
{(14,14.0) (16,14.0)};
\addplot[color=umBlue,mark=none] coordinates
{(15,15.0) (18,15.0)};
\addplot[color=umBlue,mark=none] coordinates
{(16,16.0) (19,16.0)};
\addplot[color=umBlue,mark=none] coordinates
{(17,17.0) (20,17.0)};
\addplot[color=umBlue,mark=none] coordinates
{(18,18.0) (21,18.0)};
\addplot[color=umBlue,mark=none] coordinates
{(19,19.0) (22,19.0)};
\addplot[color=umBlue,mark=none] coordinates
{(20,20.0) (23,20.0)};
\addplot[color=umBlue,mark=none] coordinates
{(21,21.0) (24,21.0)};
\addplot[color=umBlue,mark=none] coordinates
{(22,22.0) (25,22.0)};
\addplot[color=umBlue,mark=none] coordinates
{(23,23.0) (26,23.0)};
\addplot[color=umBlue,mark=none] coordinates
{(24,24.0) (27,24.0)};
\addplot[color=umBlue,mark=none] coordinates
{(25,25.0) (28,25.0)};
\addplot[color=umBlue,mark=none] coordinates
{(26,26.0) (29,26.0)};
\addplot[color=umBlue,mark=none] coordinates
{(27,27.0) (30,27.0)};
\addplot[color=umBlue,mark=none] coordinates
{(28,28.0) (31,28.0)};
\addplot[color=umBlue,mark=none] coordinates
{(29,29.0) (32,29.0)};
\addplot[color=umBlue,mark=none] coordinates
{(30,30.0) (34,30.0)};
\addplot[color=umBlue,mark=none] coordinates
{(31,31.0) (35,31.0)};
\addplot[color=umBlue,mark=none] coordinates
{(32,32.0) (36,32.0)};
\addplot[color=umBlue,mark=none] coordinates
{(34,34.0) (38,34.0)};
\addplot[color=umBlue,mark=none] coordinates
{(35,35.0) (39,35.0)};
\addplot[color=umBlue,mark=none] coordinates
{(36,36.0) (40,36.0)};
\addplot[color=umBlue,mark=none] coordinates
{(38,38.0) (42,38.0)};
\addplot[color=umBlue,mark=none] coordinates
{(39,39.0) (43,39.0)};
\addplot[color=umBlue,mark=none] coordinates
{(40,40.0) (44,40.0)};
\addplot[color=umBlue,mark=none] coordinates
{(42,42.0) (46,42.0)};
\end{axis}
\end{tikzpicture}  \label{fig:p1_2}}%
	\quad
	\subfloat[Optimal schedule, makespan = 42.]{\begin{tikzpicture}[scale=0.7]
\pgfplotsset{every axis y label/.style={
		at={(0,0.5)},
		xshift=-20pt,
		rotate=90}}
\begin{axis}[xlabel=Time, ylabel=Time,width=5.5cm,height=7cm,ymax=55,xmax=55]
\addplot[color=umBlue,mark=none] coordinates
{(0,0.0) (2,0.0)};
\addplot[color=umBlue,mark=none] coordinates
{(1,1.0) (3,1.0)};
\addplot[color=umBlue,mark=none] coordinates
{(2,2.0) (4,2.0)};
\addplot[color=umBlue,mark=none] coordinates
{(3,3.0) (5,3.0)};
\addplot[color=umBlue,mark=none] coordinates
{(4,4.0) (6,4.0)};
\addplot[color=umBlue,mark=none] coordinates
{(5,5.0) (8,5.0)};
\addplot[color=umBlue,mark=none] coordinates
{(6,6.0) (9,6.0)};
\addplot[color=umBlue,mark=none] coordinates
{(7,7.0) (10,7.0)};
\addplot[color=umBlue,mark=none] coordinates
{(8,8.0) (11,8.0)};
\addplot[color=umBlue,mark=none] coordinates
{(9,9.0) (12,9.0)};
\addplot[color=umBlue,mark=none] coordinates
{(10,10.0) (14,10.0)};
\addplot[color=umBlue,mark=none] coordinates
{(11,11.0) (13,11.0)};
\addplot[color=umBlue,mark=none] coordinates
{(12,12.0) (15,12.0)};
\addplot[color=umBlue,mark=none] coordinates
{(13,13.0) (17,13.0)};
\addplot[color=umBlue,mark=none] coordinates
{(14,14.0) (16,14.0)};
\addplot[color=umBlue,mark=none] coordinates
{(15,15.0) (18,15.0)};
\addplot[color=umBlue,mark=none] coordinates
{(16,16.0) (20,16.0)};
\addplot[color=umBlue,mark=none] coordinates
{(17,17.0) (19,17.0)};
\addplot[color=umBlue,mark=none] coordinates
{(18,18.0) (21,18.0)};
\addplot[color=umBlue,mark=none] coordinates
{(19,19.0) (23,19.0)};
\addplot[color=umBlue,mark=none] coordinates
{(20,20.0) (22,20.0)};
\addplot[color=umBlue,mark=none] coordinates
{(21,21.0) (24,21.0)};
\addplot[color=umBlue,mark=none] coordinates
{(22,22.0) (26,22.0)};
\addplot[color=umBlue,mark=none] coordinates
{(23,23.0) (25,23.0)};
\addplot[color=umBlue,mark=none] coordinates
{(24,24.0) (27,24.0)};
\addplot[color=umBlue,mark=none] coordinates
{(25,25.0) (29,25.0)};
\addplot[color=umBlue,mark=none] coordinates
{(26,26.0) (28,26.0)};
\addplot[color=umBlue,mark=none] coordinates
{(27,27.0) (30,27.0)};
\addplot[color=umBlue,mark=none] coordinates
{(28,28.0) (32,28.0)};
\addplot[color=umBlue,mark=none] coordinates
{(29,29.0) (31,29.0)};
\addplot[color=umBlue,mark=none] coordinates
{(30,30.0) (33,30.0)};
\addplot[color=umBlue,mark=none] coordinates
{(31,31.0) (35,31.0)};
\addplot[color=umBlue,mark=none] coordinates
{(32,32.0) (34,32.0)};
\addplot[color=umBlue,mark=none] coordinates
{(33,33.0) (36,33.0)};
\addplot[color=umBlue,mark=none] coordinates
{(34,34.0) (38,34.0)};
\addplot[color=umBlue,mark=none] coordinates
{(35,35.0) (37,35.0)};
\addplot[color=umBlue,mark=none] coordinates
{(36,36.0) (39,36.0)};
\addplot[color=umBlue,mark=none] coordinates
{(37,37.0) (41,37.0)};
\addplot[color=umBlue,mark=none] coordinates
{(38,38.0) (40,38.0)};
\addplot[color=umBlue,mark=none] coordinates
{(39,39.0) (42,39.0)};
\end{axis}
\end{tikzpicture}  \label{fig:p1_3}}%
	\captionsetup{width=.9\linewidth}
	\caption{Problem 1 schedules on small instance with 40 jobs}%
	\label{fig:p1}%
\end{figure}

\textbf{Job Shop Description.} 
The next scenario is a JSP  with three identical parallel machines and three job types, $a$, $b$, and $c$.
The processing times of the three job types are 2, 3, 4, and the number of jobs are 700, 700, 600 for the three job types, respectively. Minimising the makespan is this job shop reduces to makespan minimisation on parallel machines.

With this problem, an optimal plan resulting in no idle time can be created by consecutively
planning types $c, a, b$ until exhausting all type-$c$ jobs, 
and then appending this partial plan to all remaining type-$a$ and then type-$b$ jobs.
A straightforward alternative, however, is to sort all jobs by shortest processing time, thereby planning first type-$a$, then type-$b$, and lastly type-$c$ jobs.
This solution results in a $10\%$ longer makespan than the optimum and is therefore suboptimal. Figure \ref{fig:p1} shows an instance of this problem with 40 jobs where Figure \ref{fig:p1_1} depicts a random schedule, Figure \ref{fig:p1_2} the mentioned straightforward suboptimal solution and Figure \ref{fig:p1_3} the optimal solution without idle time.

\textbf{Scheduling Heuristic: Dispatching Rule for On-line Scheduling.}
The scheduling heuristic used is on-line \cite{even2009scheduling}, meaning that jobs are already processed as the schedule is being built up. 
The dispatching rule heuristic schedules the planned jobs sequentially and the jobs are started as soon as the required resources are available.
In this on-line set-up, the schedule in past time steps cannot be modified.
In this case, the known optimum described previously results in the shortest makespan.

\subsubsection{Problem 2: Demirkol Problem Instances.} \label{subsec:prob2}
The second problem case is not a specific JSP but a class of problems.
\cite{demirkol1998benchmarks} have proposed a method to generate testing instances for JSP. 
Apart from being a known benchmark for JSP, this method provides the flexibility of specifying the number of jobs, machines and several other controlling parameters.
Furthermore, it is capable of generating arbitrarily many different JSP of the same input parameters, which is ideal in this case since experiment runs are repeated.

\textbf{Job Shop Description.}
The job shop has 200 jobs.
The number of machines is 10 in order to create sufficient difficulty, as it has been shown that 6 machines are adequate to represent complex job shops \cite{raghu1993efficient}.
There are no parallel machines.
Each job has random processing times on each machine uniformly sampled from $[1, 200]$.
The order of machines that each job needs to go through is uniformly random. 
As the order is sampled from all possible permutations of machines, each job needs to be processed on each machine.
The problem definition of \cite{demirkol1998benchmarks} also includes extra parameter $T$ and $R$, which control the ratio of late jobs and the range of deadlines,  respectively.
This class of problems is generated with $T=0.2$ and $R=0.8$.
The late job ratio of $T=0.2$ is similar to those used by \cite{demirkol1998benchmarks}, but the deadline range parameter $T=0.8$ is higher than the choice of the authors ($T=0.5$).
The larger deadline range is expected to more closely resemble real-life scenarios.

\textbf{Scheduling Heuristic: Dispatching Rule for Off-line Scheduling.}
Contrary to what is used in the previous problem, the scheduling here is off-line \cite{even2009scheduling}.
This means that the complete schedule is created before being executed in the job shop.
In the dispatching rule heuristic, the planned jobs are considered sequentially, but later-considered jobs can still be scheduled to earlier time steps, since job processing only occurs after the schedule is completed.
This characteristic distinguishes the off-line set-up with its counterpart.

\subsubsection{Problem 3: Industrial Case Study} \label{subsec:industrial_case} 
The last test scenario is from a real-world industry project. It is an electro-chemical surface plating job shop with a built-in scheduling heuristic controlled by Aucos AG. Being a real-life scenario, this JSP involves multiple constraints that require additional treatment.

\textbf{Job Shop Description.}
The JSP involves 6,000 jobs of 16 types.
There are 110 machines, some of which are parallel. No buffer zones exist between the machines.
Furthermore, external resources such as racks and carriers are present and require additional modelling.
The goal is to plan thousands of jobs such that the performance of the underlying scheduling heuristic is maximised.
Since it is a real-life scenario, no optimum or bound is available.

\textbf{Additional Constraints 1: Racks.}
Prior to being processed, each job must be loaded onto a rack, and is fixed on the rack throughout the process.
Each type of job can only be processed on one type of rack, but one type of rack can be used for different types of jobs.
As the behaviour of the scheduling heuristic is not known,
the order of the completed jobs and therefore the availability of racks is not known for certainty.
However, as the number of racks is limited, missing the correct racks would cause delay.
In this context, the availability of racks must be considered when searching for plans.
In total, there are 303 types of racks. Each type has 3 racks on average but there are a few dominant rack types having approximately 20 racks.

\textbf{Additional Constraints 2: Carriers.}
The racks with mounted jobs have to be attached to carriers to move through the line.
The planning task therefore is decomposed to the assignment of jobs to racks and racks to carriers.
As there are fewer racks than jobs and fewer carriers than racks, it is sometimes necessary to exchange rack on the carriers with other types such that jobs can be further processed.
This action is called rack change.
As rack changes also need to be planned, they are considered as a special job type taking constant time.
There are 80 carriers in total.

\textbf{Additional Constraints 3: Loading Area.}
Another set of constraints is caused by the presence of the loading area. In the preparatory area of this line, jobs are loaded onto racks at 6 parallel stations. 
This means that job loading is parallelised.
Since the line runs cyclically, the loaded jobs are taken from the loading area into the production plant in every fixed time interval.
This action of taking loaded jobs into the line is the loading process.
Due to mechanical constraints, only two loading processes can be completed during every time interval.
The loading process is modelled as a multiple server polling system \cite{borst1998waiting}.
In the polling system, the waiting lines represent the jobs at the 6 loading stations.
The inter-arrival time is the time taken to load the job.
As the loading time differs from job to job, the inter-arrival time is different between queues and changes within the queue.
Two polling servers represent the job loader.
In general, multiple server polling systems are hard to solve \cite{borst1998waiting, van1997analysis} and no solution for such systems with non-constant inter-arrival times has been found.
In this work, a simulation was made to model this polling system. 
The simulation alone is too slow to use during the search. 
Therefore, a tabling strategy was used, ensuring that the same situation is only evaluated once.
This involves a lookup table with large number of combinations of jobs on loading stations.
Simulation is only made if there is no table entry.

\subsection{Baselines and Benchmarks}\label{sec:benchmark}
The baselines consist of several scheduling heuristics, including Longest Processing Time (LPT), Shortest Processing Time (SPT) \cite{holthaus1997efficient}, as well as Earliest Due Date (EDD) \cite{demirkol1998benchmarks}. Since they are relatively unsophisticated and domain-dependent, the H-MCTS is expected to outperform them.

As benchmark five types of Monte-Carlo techniques for planning are tested, namely flat Monte-Carlo Search (i.e. a 1-ply deep Monte-Carlo Search, denoted as MCS), Nested Monte-Carlo Search (NMCS), its hierarchical variant (H-NMCS), MCTS, and H-MCTS. Moreover, the effect of MCTS search enhancements are tested.
The combination of enhancements that provided the best results is utilised in further experiments for abstractions, objective modification, and the industrial case study. 
The benchmarks are known optima or bounds, which H-MCTS should be as close to as possible. For problems where finding the optima is too time consuming, lower bounds of the makespan is derived using integer linear programming (ILP). With the industrial case study, due to the encapsulated heuristic, there is no known optimum or bound. In this case, the performance of the proposed approach is compared with the original situations in the plant prior to the implementation of the new planning system.

\subsection{Reported Metrics}\label{sec:metric}
In the initial experiment regarding learning the scheduling heuristic, the accuracy compared to the ground truth is reported. In all further experiments regarding the search performance, the evaluation metrics are based makespan or lateness, which are both functions of the job completion time.
For problems with known optima, the ratios between the achieved results and the optima are reported. When the optima are unknown, the ratios to some lower bound are used instead. In either case, a value closer to $1$ indicates better performance.  Unless otherwise mentioned, the reported metrics are averaged from 10 randomly initialised runs with $95\%$-confidence intervals. The standard deviations are also provided.

%%%%%%%%%%%%%%%%%%%%%%%%%%%%%%%%%%%%%%%%%%%%%%%%%%%%%%%%%%%%%%%%%%%%%%%%%%%%%%%%%
%%%   Results
%%%%%%%%%%%%%%%%%%%%%%%%%%%%%%%%%%%%%%%%%%%%%%%%%%%%%%%%%%%%%%%%%%%%%%%%%%%%%%%%%

\section{Results}
\label{sec:results}
Experiments on heuristic learning, different Monte-Carlo techniques, abstraction approaches, modified objective and a case study are presented and discussed in Subsections \ref{sec:expMCS} to \ref{sec:expCaseStudy}, respectively. All experiments are executed on a standard desktop quad-core processor architecture with hyper-threading from year 2018 using the MSVC2017 compiler in the Qt framework.

%% DO NOT PUT BACK
%%\subsection{Heuristic Learning Results}\label{sec:expHeuLearn} 
%%\input{Chapter4/HeuristicLearn.tex}

\subsection{Search Results} \label{sec:expMCS}

The experiments in this section aim to compare the performance of H-MCTS with different baselines and Monte-Carlo search variants, as well as the impact of different enhancements on H-MCTS.

\subsubsection{Monte-Carlo Search Comparisions}
Five Monte-Carlo techniques, namely flat MCS, MCTS, NMCS, H-MCTS, H-NMCS are tested on Problem 1. This problem case is chosen because of its simple structure and known optimum, which allows qualitative analyses of the behaviour of the tested techniques.

Besides H-MCTS and H-NMCS using  integrated abstraction, all variants are run without any other enhancements or parallelisation. The total computation time is held equal for all techniques (i.e., 10 seconds per search step, see \ref{subsec:adjtime}). NMCS is executed with nesting level 2.  No higher nesting level is used due to the computational load of the exponentially-growing number of evaluations. 
\begin{table}[H]
	\centering
		\begin{tabular}{@{}lcc@{}}
		\toprule
			Method & Ratio to ILP bound & Standard deviation \\
		\midrule
			Optimum & 1.000 $\pm$0.000 & - \\
			SPT-Heuristic & 1.100 $\pm$0.000 & - \\
			Random                 & 1.324 $\pm$0.019& 0.010 \\
			Flat MCS              & 1.191 $\pm$0.014& 0.007 \\
			MCTS                    & 1.184 $\pm$0.021& 0.011 \\
			NMCS                    & 1.185 $\pm$0.008& 0.004 \\
			H-MCTS & 1.031 $\pm$0.019& 0.010 \\
			H-NMCS & 1.146 $\pm$0.016& 0.008 \\
		\bottomrule
	\end{tabular}

\iffalse

\begin{tikzpicture}
\begin{axis}[
ylabel=Ratio to optimum,
xtick={0,1,2,3,4,5,6,7},
xticklabels={Optimum,SPF,Random,Flat MCS,MCTS,NMCS,Hierarchical MCTS, Hierarchical NMCS},
x tick label style={rotate=90,anchor=east}
]
\addplot[
%smooth,
only marks,
mark=*,
blue,
error bars/.cd, y dir=both, y explicit,
] plot coordinates {
	(0,1)
	(1,1.1)
	(2,1.324) +=(0,0.019) -= (0,0.019)
	(3,1.191) +=(0,0.014) -= (0,0.014)
	(4,1.184) +=(0,0.021) -= (0,0.021)
	(5,1.185) +=(0,0.008) -= (0,0.008)
	(6,1.031) +=(0,0.019) -= (0,0.019)
	(7,1.146) +=(0,0.016) -= (0,0.016)
};
\end{axis}
\end{tikzpicture}
\fi 
	\captionsetup{width=.9\linewidth}
	\caption{Evaluation of different search techniques on Problem 1 with 2000 jobs.}
	\label{tab:mcts_methods}
\end{table}

In Table \ref{tab:mcts_methods}, the first row is the optimum with no idle time.
The second row corresponds to the SPT heuristic, which plans jobs by ascending processing time.
The schedule created by SPT is suboptimal solution described in Subsection \ref{subsec:prob1}.
Nevertheless, it has to be noted that, despite being suboptimal, the solution found by SPT is rather strong, 
since the design of this heuristic is suited to the specific problem.
No significant difference can be observed among results of flat MCS, MCTS, NMCS and H-NMCS. 
Their performance does not surpass the SPT baseline.

H-MCTS, however, significantly outperforms the other techniques, including the SPT baseline.
The schedules it creates closely resemble the optimum but have occasional deviations.
This observation signifies that H-MCTS is capable of distinguishing more promising regions in the search space due to the hierarchy.
Interestingly, another hierarchical technique, H-NMCS, does not exhibit equally strong performance as H-MCTS.
This difference can be attributed to two possible reasons. 
Firstly, the nesting level in this experiment is limited at 2 due to the exponential growth of evaluations.
Secondly, H-NMCS performs no exploration beyond its nesting level.
The explorations that it makes within the nesting levels do not distinguish the child nodes based on their qualities. 

\begin{figure}[H]%
	\centering
	\subfloat[Flat MCS search tree.]{\begin{tikzpicture}
\pgfplotsset{every axis y label/.style={
		at={(0,0.5)},
		xshift=-10pt,
		rotate=90}}
\pgfplotsset{every axis x label/.style={
		at={(0.5,0)},
		below,
		yshift=-5pt}}
\pgfplotsset{ticks=none}
\begin{axis}[enlargelimits=false, axis on top, axis equal image, width=0.55\textwidth,ylabel=Branches,xlabel=Jobs,axis lines=left, xtick=\empty, ytick=\empty]
\addplot graphics [xmin=0,xmax=100,ymin=0,ymax=59] {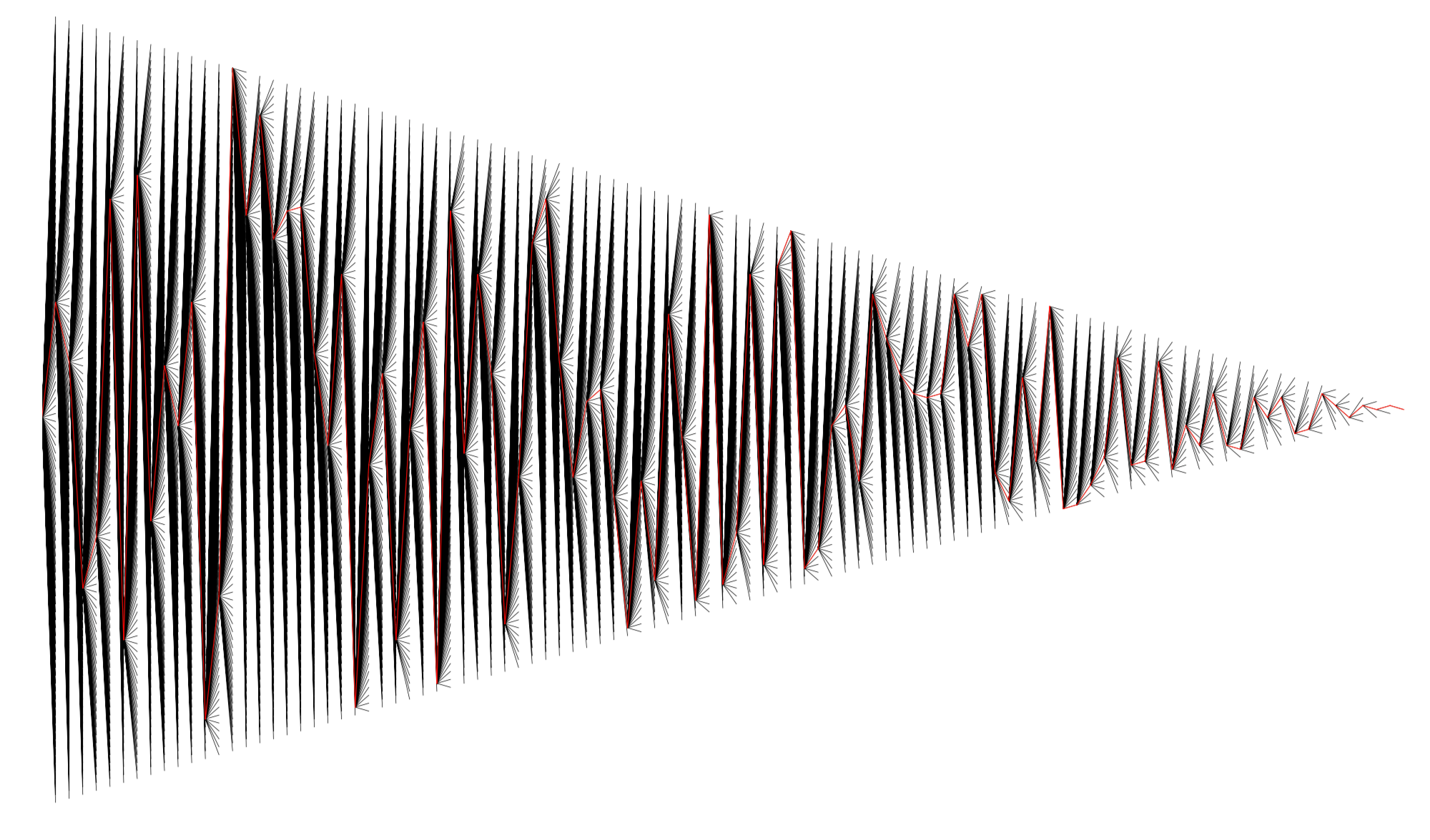};

\end{axis}
\end{tikzpicture}

 	   \label{fig:t_flat}} 
	\subfloat[H-NMCS search tree level 2.]{\begin{tikzpicture}
\pgfplotsset{every axis y label/.style={
		at={(0,0.5)},
		xshift=-10pt,
		rotate=90}}
\pgfplotsset{every axis x label/.style={
		at={(0.5,0)},
		below,
		yshift=-5pt}}
\pgfplotsset{ticks=none}
\begin{axis}[enlargelimits=false, axis on top, axis equal image, width=0.55\textwidth,ylabel=Branches,xlabel=Jobs,axis lines=left, xtick=\empty, ytick=\empty]
\addplot graphics [xmin=0,xmax=100,ymin=0,ymax=59] {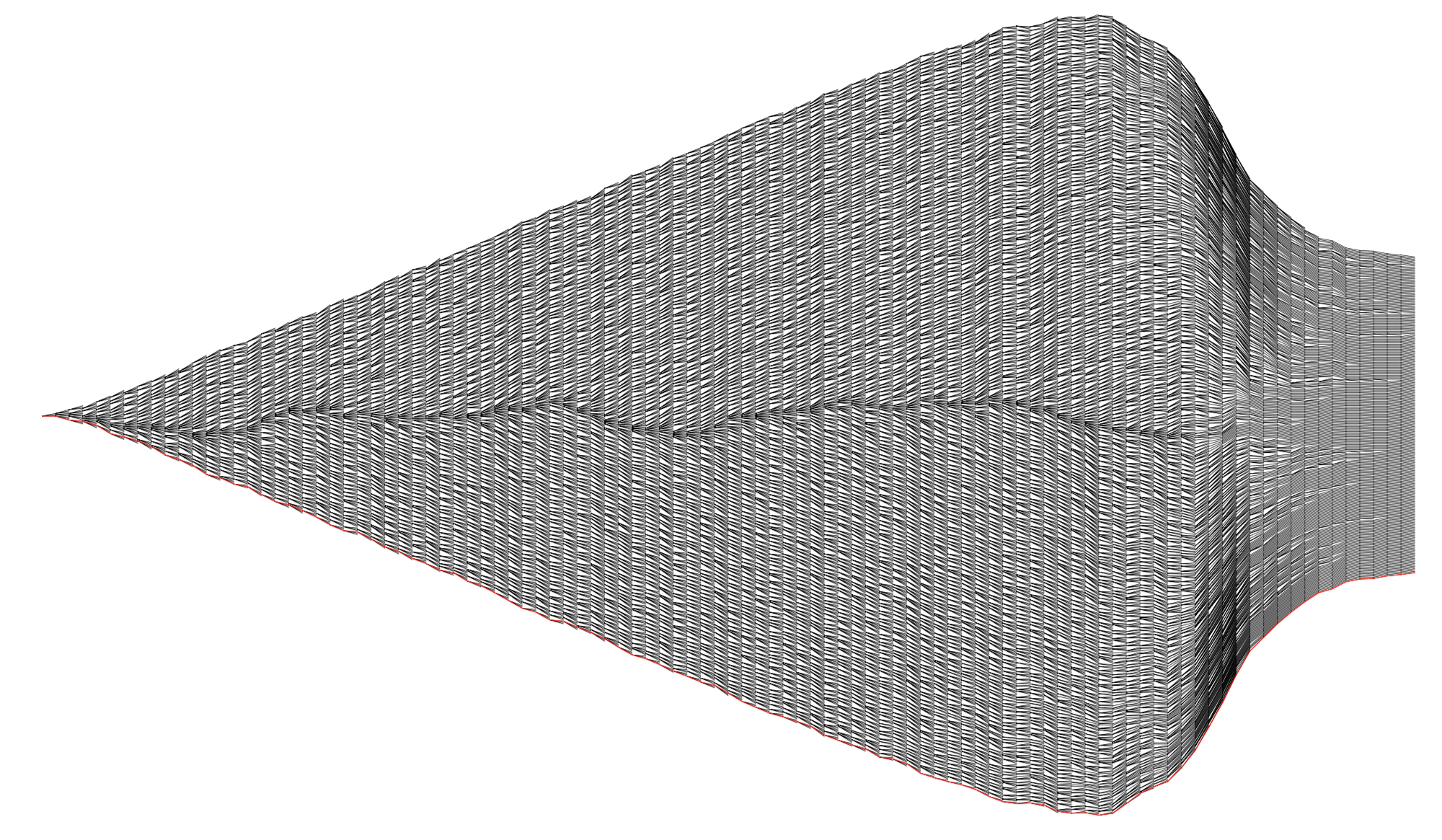};

\end{axis}
\end{tikzpicture}

 	   \label{fig:t_nested}} \\
	\subfloat[MCTS search tree.]{\begin{tikzpicture}
\pgfplotsset{every axis y label/.style={
		at={(0,0.5)},
		xshift=-10pt,
		rotate=90}}
\pgfplotsset{every axis x label/.style={
		at={(0.5,0)},
		below,
		yshift=-5pt}}
\pgfplotsset{ticks=none}
\begin{axis}[enlargelimits=false, axis on top, axis equal image, width=0.55\textwidth,ylabel=Branches,xlabel=Jobs,axis lines=left, xtick=\empty, ytick=\empty]
\addplot graphics [xmin=0,xmax=100,ymin=0,ymax=59] {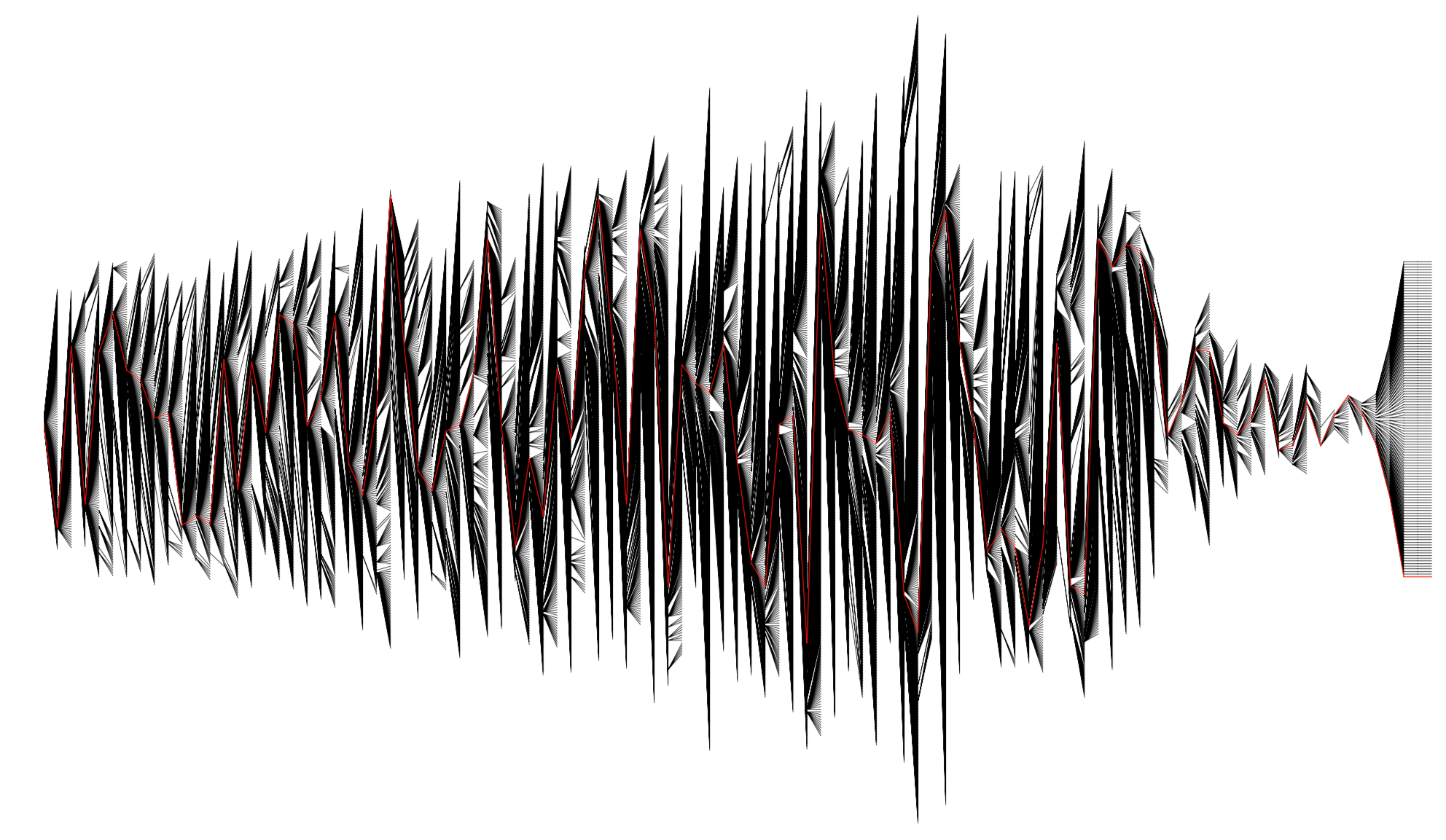};

\end{axis}
\end{tikzpicture}

 	   \label{fig:t_mcts}} 
	\subfloat[H-MCTS search tree.]{\begin{tikzpicture}
\pgfplotsset{every axis y label/.style={
		at={(0,0.5)},
		xshift=-10pt,
		rotate=90}}
\pgfplotsset{every axis x label/.style={
		at={(0.5,0)},
		below,
		yshift=-5pt}}
\pgfplotsset{ticks=none}
\begin{axis}[enlargelimits=false, axis on top, axis equal image, width=0.55\textwidth,ylabel=Branches,xlabel=Jobs,axis lines=left, xtick=\empty, ytick=\empty]
\addplot graphics [xmin=0,xmax=100,ymin=0,ymax=59] {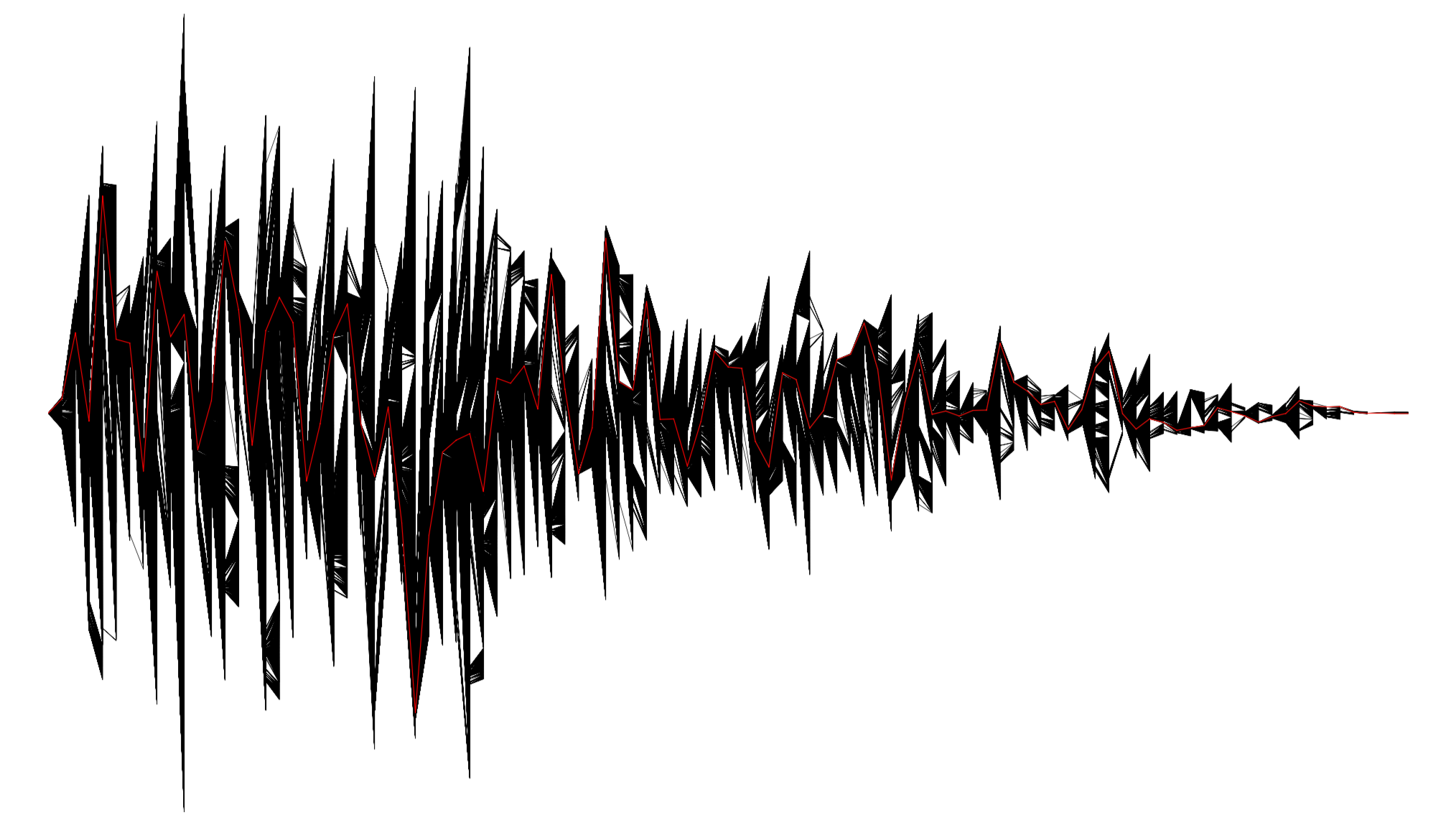};

\end{axis}
\end{tikzpicture}

 	   \label{fig:t_hierarchy}} \\
	\captionsetup{width=.9\linewidth}
	\caption{Search trees of different Monte-Carlo techniques generated on Problem 1. The number of jobs is reduced to 100 for visual clarity.}%
	\label{fig:treePlots}%
\end{figure}

An additional qualitative analysis is performed by visualising the search trees.
Figure \ref{fig:treePlots} shows the search trees of flat MCS, H-NMCS, MCTS, and H-MCTS. 
Here a reduced version of Problem 1 consisting of 100 jobs is used for clear visualisation.

From Figure \ref{fig:t_flat}, it can be seen that flat MCS does not expand nodes selectively. 
Instead, the best child from uniform samples is chosen at each step, resulting in the shrinking number child nodes as the search progresses.
This causes the triangular shape with reducing number of jobs towards the end of the search.
This observation further substantiates the time-redistribution enhancement introduced in Subsection \ref{subsec:enhancements}, as the search time is distributed according to the shape of the flat MCS.
Figure \ref{fig:t_nested} depicts the search tree of H-NMCS, whose width growth is nearly anti-proportional to that of flat MCS.
The tree width grows exponentially with the nesting level, as every node is followed by another nested MCS.
With nesting level 1, the resulting tree would be same as that of flat MCS. 
In Figure \ref{fig:t_mcts} for MCTS, the tree widens as the search progresses. 
As the search progresses, the number of children at one node get less and therefore, the tree is expanded deeper and thus wider. 
When approaching the terminal nodes, the tree width reduces again, as the number of remaining jobs is sufficiently low and it is possible to enumerate all child nodes.
Lastly, Figure \ref{fig:t_hierarchy} visualises the search tree of H-MCTS, which is clearly more focused, due to the hierarchical structure.

\subsubsection{Monte-Carlo Tree Search Enhancements} \label{subsec:enhancements_exp}
The following series of experiments test the MCTS enhancements described in Subsection \ref{subsec:enhancements}. 
The aforementioned root and tree parallelisation each uses 4 threads. As test case Problem 2 is chosen as it is more complex than Problem 1. It is expected that the effects of the enhancements are more visible on more challenging problems.

Two benchmarks are used, namely the trivial lower bound and strong ILP bound. 
The former is simply the maximum of the longest job and the busy time of the most occupied machine.
Despite its simplicity, this bound is a commonly used in the literature \cite{williamson1997short}.
%%The latter is derived by the ILP formulation in Figure \ref{fig:ilp_makespan} in Appendix \ref{app:ilp_c}.
Due to long computation time, the ILP solver is not run until finding the optimum. Instead, the solution is taken when the gap is reduced to 30\% and no changes occur in the incumbent solution for multiple epochs. The ILP solution is used as a strong lower bound on the makespan.
Additionally, two baseline heuristics, namely SPT and LPT, are included in the comparison, as they have shown good performance in minimising makespan \cite{holthaus1997efficient}.

\begin{table}[H]
	\centering
	\input{figures/mcts_enhance} 
	\captionsetup{width=.9\linewidth}
	\caption{Evaluation of different enhancement on 5 instances of Problem 2 with 200 jobs. Objective is minimising makespan.}
	\label{tab:mcts_enhance}
\end{table}

Table \ref{tab:mcts_enhance} outlines the full comparison based on 5 problem instances with 200 jobs. 
In each configuration, the average results from 10 randomly initialised runs are reported.
Basic MCTS without any enhancement already significantly outperforms SPT and LPT.
This shows that on complex problems lacking structure, conventional heuristics cannot perform as well as MCTS.
Moreover, time re-distribution provides a significant performance boost from the basic MCTS.
The effects of adding best path memorisation and progressive history are not evident.
Since the problem instances are generated with many random parameters, 
there may be too little structured past knowledge for these two enhancements to take effect.
The best performance is achieved by incorporating all enhancements as well parallelisation, reaching a ratio of 1.159 to the strong ILP bound.

To validate the stability of the search performance, in the next experiment, one single search run is conducted on a larger collection of 50 problem instances, with the results shown in Table \ref{tab:mcts_enhance50}.
This set-up contrasts the previous one, which involves only 5 instances and aggregates results from 10 runs.
Note that only the fully enhanced MCTS is tested, as the goal of this experiment is to investigate search stability instead of comparing the individual enhancements.
It can be observed that the strength of the enhanced MCTS remains, outperforming both baseline heuristics significantly.

\begin{table}[H]
	\centering
	\begin{tabular}{@{}lcc@{}}
		\toprule
		Method & Ratio to lower bound & Std. dev. \\
		\midrule
		SPT-Heuristic&	1.727 $\pm$0.176 & 0.090 \\
		LPT-Heuristic&	1.458 $\pm$0.125 & 0.064 \\
		Random&	1.667	$\pm$0.365 & 0.186 \\
		%TR + BP + PH + Parallel& 1.159 $\pm$0.047 & 0.024 \\ copied wrong row from excel
		TR + BP + PH + Parallel& 1.233 $\pm$0.042 & 0.021 \\
		\bottomrule
	\end{tabular} 
	\captionsetup{width=.9\linewidth}
	\caption{Evaluation of different enhancement on 50 instances of Problem 2 with 200 jobs but only one search run. Objective is minimising makespan.}
	\label{tab:mcts_enhance50}
\end{table}

%-----------------------------qualitative-----------------------------------
\begin{figure}[H]%
	\centering
	\subfloat[SPT Heuristic (1.7 opt)]{\includegraphics[width=0.45\textwidth]{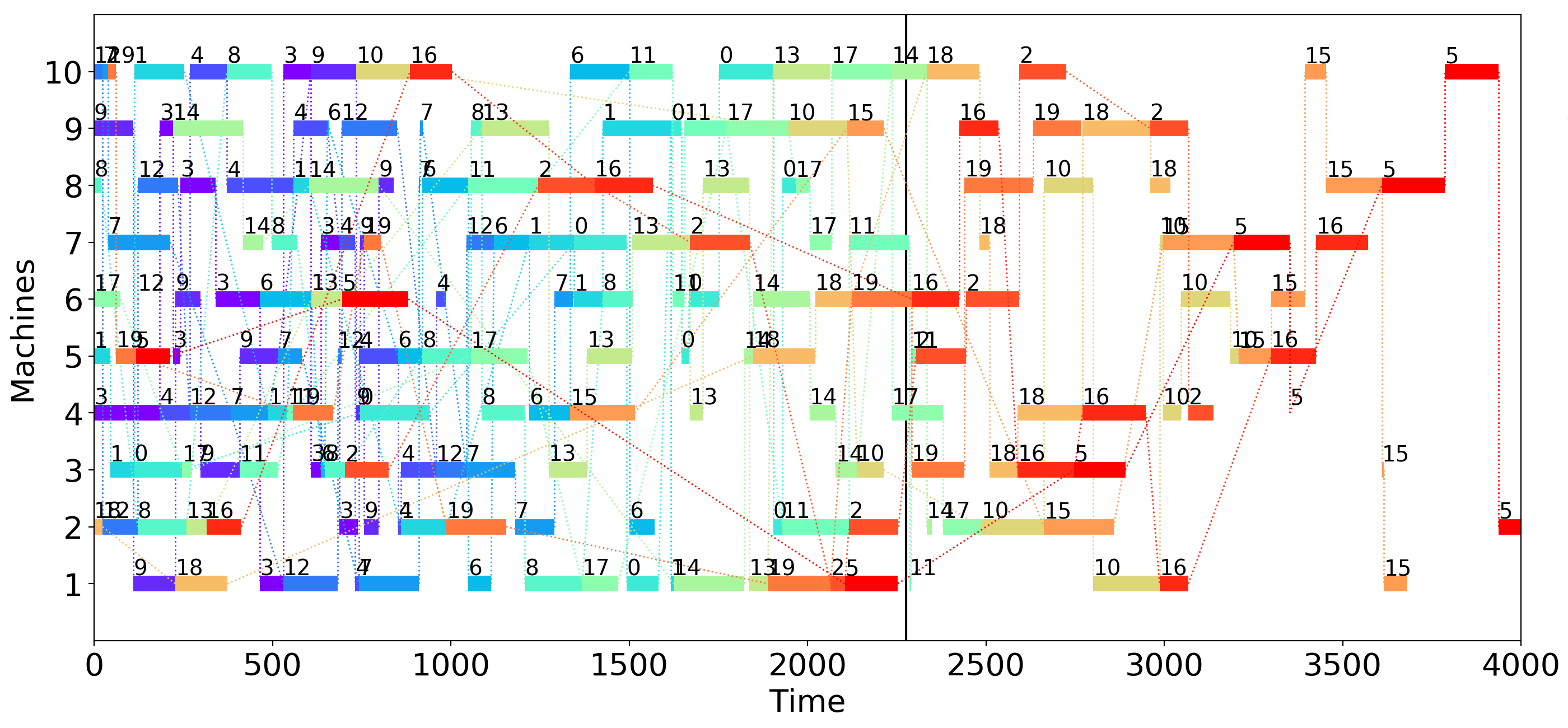}  \label{fig:s_spt}} 
	\subfloat[LPT Heuristic (1.4 opt)]{\includegraphics[width=0.45\textwidth]{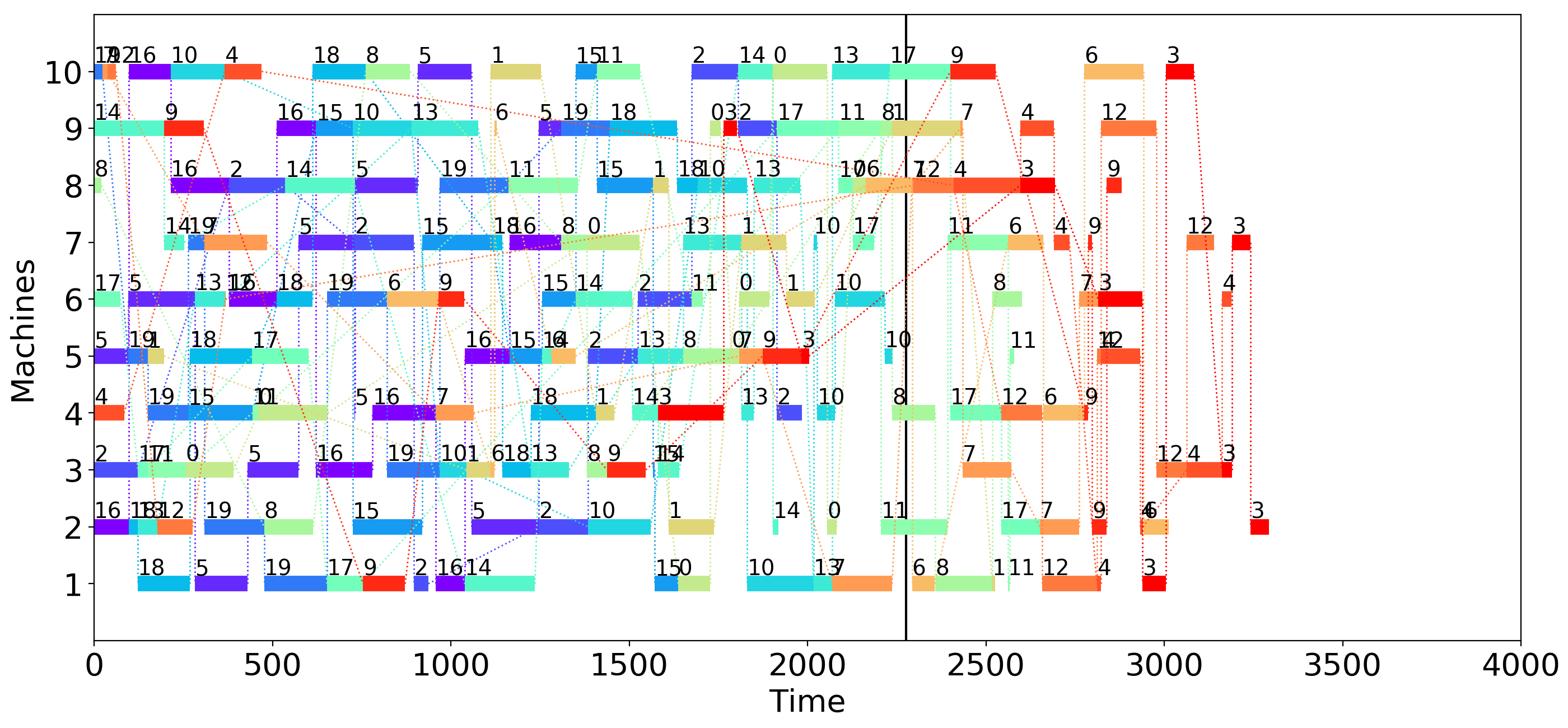}  \label{fig:s_lpt}} \\
	\subfloat[Random order (1.5 opt)]{\includegraphics[width=0.45\textwidth]{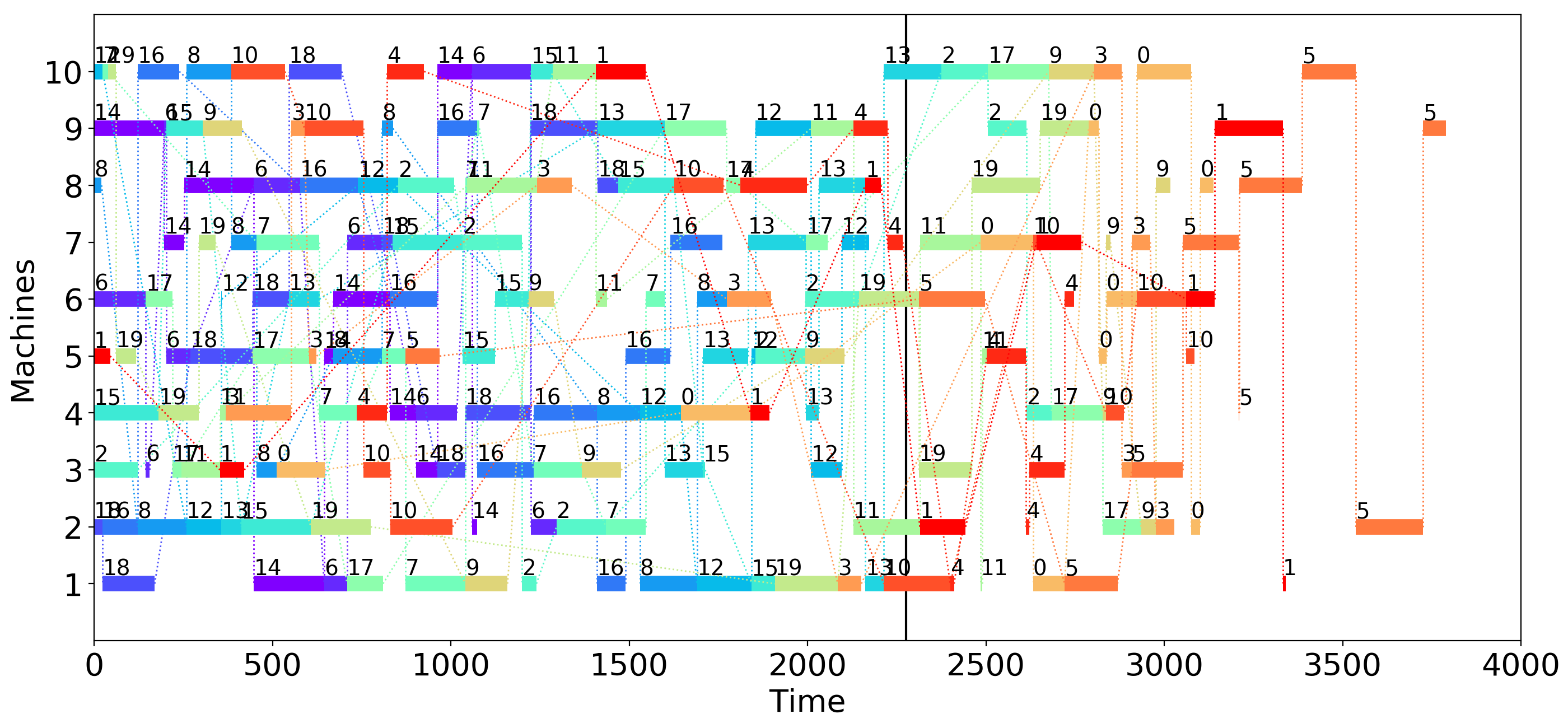}  \label{fig:s_rand}} 
	\subfloat[Optimal schedule (1 opt)]{\includegraphics[width=0.45\textwidth]{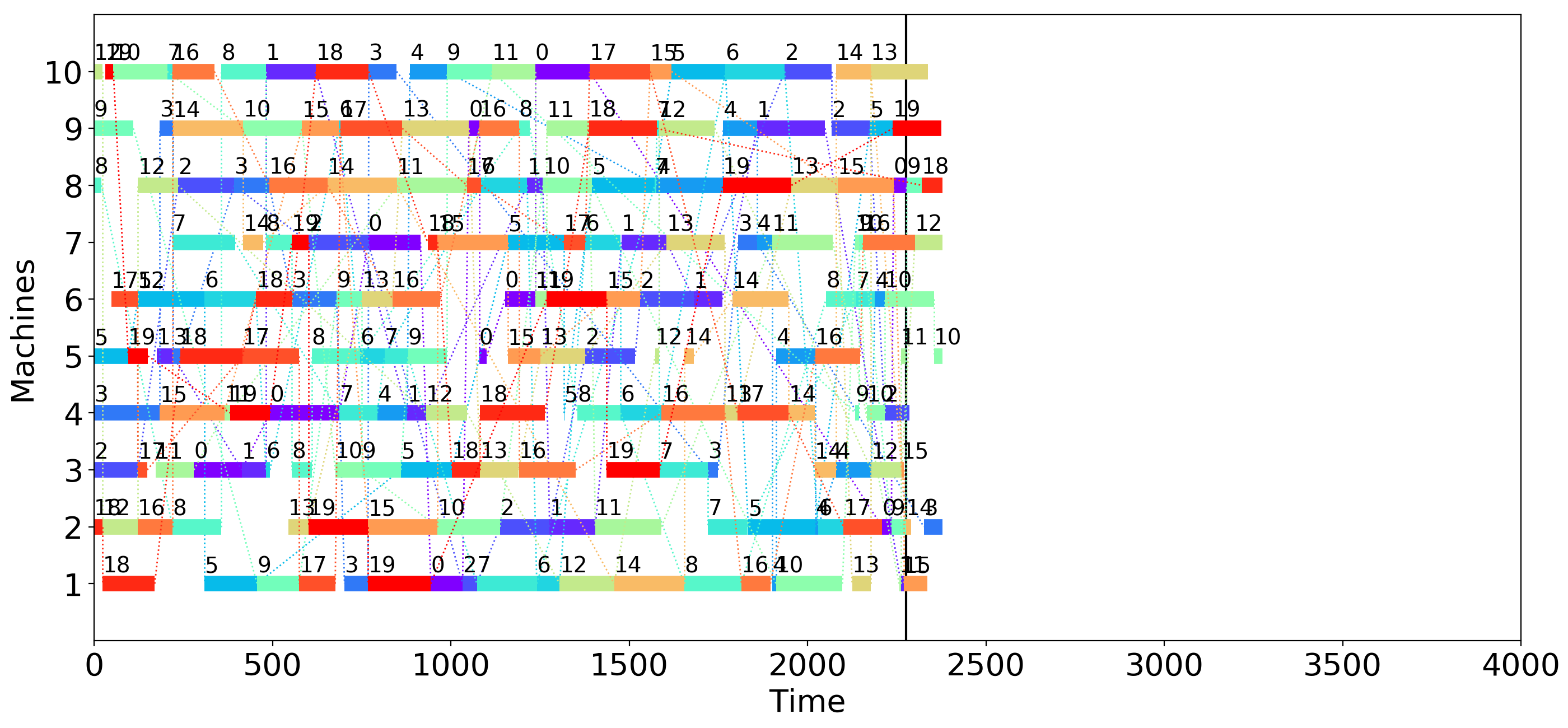}  \label{fig:s_opt}} \\
	\subfloat[MCTS (1.2 opt)]{\includegraphics[width=0.45\textwidth]{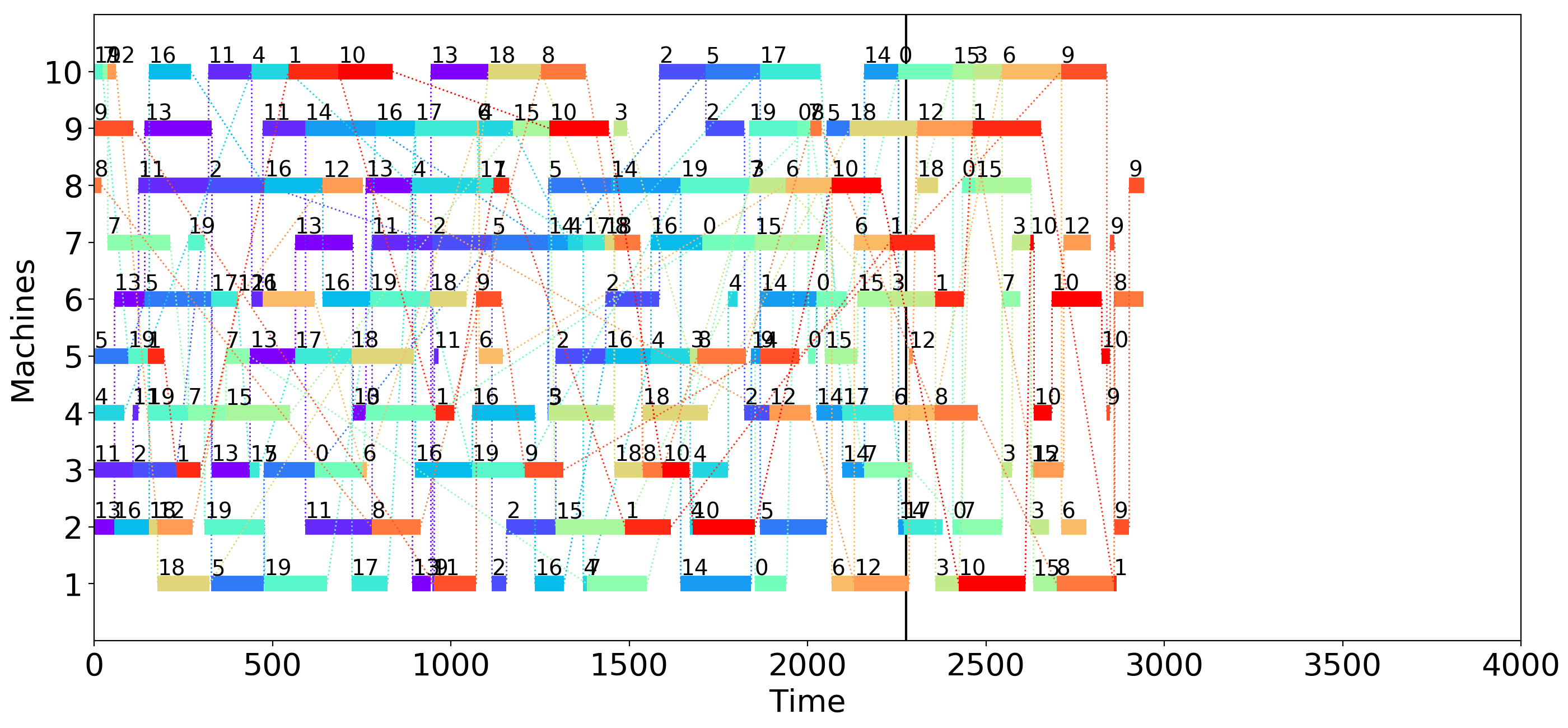}  \label{fig:s_mcts}}
	\subfloat[Enhanced MCTS (1.1 opt)]{\includegraphics[width=0.45\textwidth]{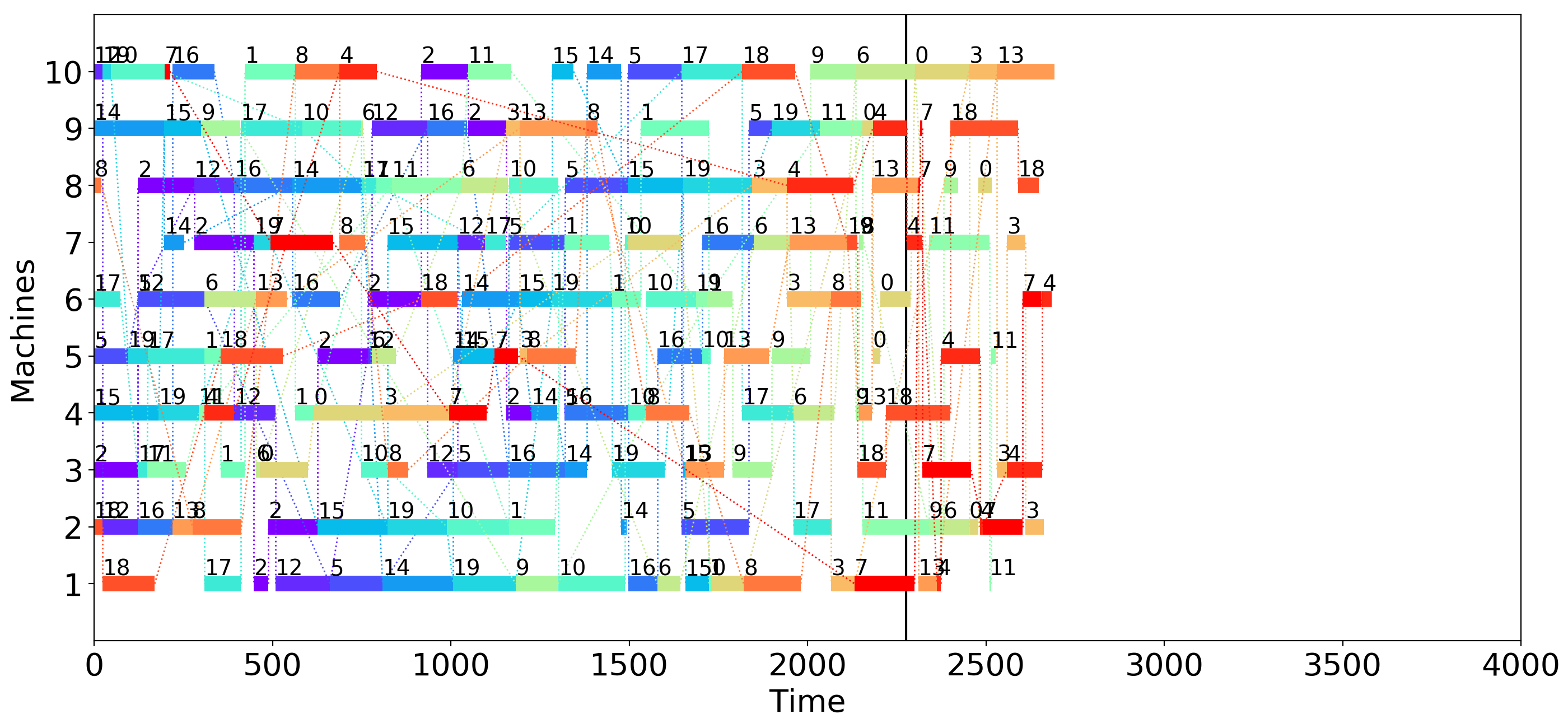}  \label{fig:s_enhanced}} %
	\captionsetup{width=.9\linewidth}
	\caption{Resulting schedules of 20 jobs of Problem 2 with minimising makespan as objective. Subfigures (a) to (f) are for reference and (e) and (f) are results after planning with different MCTS techniques. The vertical black line represents the trivial lower bound of the makespan, which is the maximum of the longest job and the sum of occupation time on the most used machine. A full description of the jobs is given in \cite{wimmenauer2019}. 
	}
	\label{fig:schedules}%
\end{figure}

Figure \ref{fig:schedules} shows the schedules on a smaller instance of 20 jobs on 10 machines.
Detailed parameters of this problem can be found in \cite{wimmenauer2019}.
As the objective is to minimise makespan, a schedule is better if its components are visually closer.
A gap indicates idle time on a machine.
It can be seen that LPT in Figure \ref{fig:s_lpt} creates a denser schedule than SPT in Figure \ref{fig:s_spt}.
Following the longest-processing-first rule, LPT schedules jobs with longer processing time first and subsequently fitting shorter jobs between the gaps between the longer jobs.
Intuitively, this creates more compact schedules than SPT, which follows the reversed rule.
It is also visible that MCTS in Figure \ref{fig:s_mcts} creates denser schedules than both LPT and SPT.
A further improvement can be seen in Figure \ref{fig:s_enhanced} with all enhancements.

\subsection{Abstraction Results}\label{sec:expAbstraction}

The experiments in this section investigate the effects of abstraction.
The detached and integrated abstraction described in Subsection \ref{subsec:jobAbstraction} are evaluated. Both approaches are compared to the strong ILP bound and the basic MCTS without abstraction. 

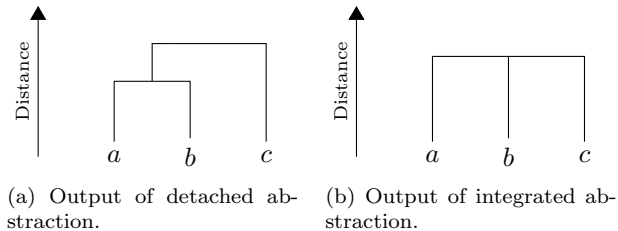
\begin{figure}[!ht]%
	\centering
	\subfloat[Output of detached abstraction.]{\begin{tikzpicture}[sloped]
\node (a) at (2,0) {$a$};
\node (b) at (3,0) {$b$};
\node (c) at (4,0) {$c$};
\node (ab) at (2.5,1) {};
\node (abc) at (2.75,1.5) {};

\draw  (a) |- (ab.center);
\draw  (b) |- (ab.center);
\draw  (c) |- (abc.center);
\draw  (ab.center) |- (abc.center);

\draw[->,-triangle 60] (1,0) -- node[above]{\scriptsize Distance} (1,2);
\end{tikzpicture} \label{fig:e_h1}}%
	\quad
	\subfloat[Output of integrated abstraction.]{\begin{tikzpicture}[sloped]
\node (a) at (2,0) {$a$};
\node (b) at (3,0) {$b$};
\node (c) at (4,0) {$c$};
\node (abc) at (3,1.33) {};

\draw  (a) |- (abc.center);
\draw  (b) |- (abc.center);
\draw  (c) |- (abc.center);

\draw[->,-triangle 60] (1,0) -- node[above]{\scriptsize Distance} (1,2);
\end{tikzpicture}  \label{fig:e_h3}}%
	\captionsetup{width=.9\linewidth}
	\caption{Output of two abstraction approaches on three jobs where the pair-wise distances are the same.}%
	\label{fig:e_hierarchy}%
\end{figure}

Firstly, the effect of abstraction is tested in a base scenario where no search enhancement is involved.
The first half of Table \ref{tab:abstract} shows that the integrated abstraction significantly outperforms the detached version for Problem 1.
An inspection of the output reveals that the abstractions constructed by the two approaches are different as shown in Figure \ref{fig:e_hierarchy}.

\begin{table}[H]
	\centering
	\begin{tabular}{@{}lcccc@{}}
		\toprule
		& \multicolumn{2}{c}{Problem 1} & \multicolumn{2}{c}{Problem 2} \\
		\cmidrule(lr){2-3} \cmidrule(lr){4-5} 
		Method & Optimum & Std.dev. & Ratio to ILP & Std.dev.\\
		\midrule
		Best solution & 1.000 $\pm$0.000 & - & 1.000 $\pm$0.008& 0.004 \\
		No abstraction                     & 1.184 $\pm$0.022& 0.011 & 1.158 $\pm$0.028 & 0.014\\
		Detached abstraction  & 1.143 $\pm$0.032& 0.016 & 1.163 $\pm$0.031 & 0.016 \\
		Integrated abstraction  & 1.031 $\pm$0.019& 0.010 & 1.158 $\pm$0.029 & 0.015\\
		\bottomrule
	\end{tabular} 
	\captionsetup{width=.9\linewidth}
	\caption{Evaluation of different abstraction approaches on Problems 1 and 2. No enhancement is used in any of the experimented configurations.}
	\label{tab:abstract}
\end{table}

On the one hand, it is clear from Figure \ref{fig:e_h1} that, as hierarchical clustering is sensitive to input orders, the detached abstraction created an imbalanced hierarchy although the pairwise distances are the same.
On the other hand, integrating the abstraction in the search creates a more balanced hierarchy, as shown in Figure \ref{fig:e_h3}.
This observation provides insight regarding the performance differences.
That is, a more balanced abstraction hierarchy leads to more uniform sampling over the jobs.
In an unbalanced hierarchy, one abstraction branch has fewer jobs than other branches, for example $c$ in Figure \ref{fig:e_h1}, sampling will provide better estimates of this branch than others.

It can be seen from the right half of Table \ref{tab:abstract} that no performance gain is achieved when applying abstraction on Problem 2.
This is expected, however, as the parameters of the jobs are randomly generated. As the job features are uniformly random, the pairwise distances among jobs are all similar. This undermines the basis of the distance-based abstraction approach.  Moreover, from a theoretical point of view, without distinguishable features, abstraction is no longer meaningful.

Now that the performance gain of integrated abstraction is shown in the base scenario, a further experiment is conducted with all search enhancements incorporated on Problem 1, and compared to the ones without the search enhancements as previously given in Table \ref{tab:mcts_methods}.  Note that as the detached abstraction approach has exposed its drawback of unbalanced hierarchy from the previous experiment, it is excluded in this comparison.
Table \ref{tab:abstractEnhance} provides the results of different configurations. 
Based on the basic configuration with neither enhancement nor abstraction, adding enhancements provides a minor performance gain, improving the ratio to optimum from 1.184 to 1.103.
The integrated abstraction, though, provides a much larger gain, further reducing the ratio to 1.031. 
This can be attributed to the fact that abstraction is able to better exploit the internal structure of the problem.
Moreover, the strongest configuration is with both abstraction and search enhancements, reaching a new low ratio to optimum of 1.022.
This showcases that search enhancements and abstractions are both valuable additions that can achieve greater positive effects when combined.

\begin{table}[H]
	\centering
	\begin{tabular}{@{}lcccc@{}}
		\toprule
		& \multicolumn{2}{c}{(H-)MCTS} & \multicolumn{2}{c}{Enhanced (H-)MCTS} \\
		\cmidrule(lr){2-3} \cmidrule(lr){4-5} 
		Method & Optimum & Std.dev. & Optimum & Std.dev.\\
		\midrule
		Optimum & 1.000 $\pm$0.000 & - & 1.000 $\pm$0.000& - \\
		No abstraction                     & 1.184 $\pm$0.022& 0.011 & 1.103 $\pm$0.020 & 0.010\\
		Integrated abstraction  & 1.031 $\pm$0.019& 0.010 & 1.022 $\pm$0.016 & 0.008\\
		\bottomrule
	\end{tabular}

%1-(1,158/1,247) = 0,0714
%1-(1,021/1,031) = 0,0087 
	\captionsetup{width=.9\linewidth}
	\caption{Evaluation of integrated abstraction based on MCTS with and without enhancements on Problem 1.}
	\label{tab:abstractEnhance}
\end{table}

%%%%%%%%%%%%%%%%%%%%%%%%%%%%%%%%%%%%%%%%%%%%%%%%%%%%%%%%%%%%%%%%%%%%%%%%%%%%%%%%%

\subsection{Optimizing for Different Objective} \label{sec:expObjective}

We also investigate to which extent we can optimize for a different objective, while naturally not changing the underlying blackbox scheduling heuristic. We experiment with modifications to the evaluation metric used in MCTS.
As the search technique we use the enhanced MCTS without abstraction.

%%\subsection{Minimising Maximum Lateness} \label{subsec:min_max_lateness}
Concretely, we set the optimization goal to minimize the maximum lateness.
For a more complete comparison, a heuristic designed for this new objective is added as baseline. 
This is the earliest due-dates (EDD) \cite{holthaus1997efficient} heuristic.
Again a strong ILP bound is created under the same conditions as 
specified before in Subsection \ref{subsec:enhancements_exp}.
%%The detailed ILP formulation can be found in Figure \ref{fig:ilp_lateness} in Appendix \ref{app:ilp_l}.
Previously used SPT and LPT are included as baselines. Although they are not designed for minimising maximum lateness, investigating their performance under the new situation may provide additional insights.
For the search, the only modification needed is the evaluation metric.

\begin{table}[H]
	\centering
		\begin{tabular}{@{}lcc@{}}
		\toprule
		& Ratio to &  \\
		Method & ILP bound & Std. dev. \\
		\midrule
		Strong ILP bound & 1.000 $\pm$0.000& - \\
		EDD-Heuristic                 & 2.253 $\pm$0.586 & 0.299\\ %4189
		SPT-Heuristic                 & 2.978 $\pm$0.719& 0.367\\ %4189
		LPT-Heuristic                 & 2.865 $\pm$0.790& 0.403\\ %4189
		Random                 & 3.134 $\pm$1.623&  0.828 \\
		Enhanced MCTS & 1.781 $\pm$0.455& 0.232 \\
		\bottomrule
	\end{tabular} 
	\captionsetup{width=.9\linewidth}
	\caption{Evaluation of minimizing maximum lateness on Problem 2 with 200 jobs.}
	\label{tab:lateness}
\end{table}

Table \ref{tab:lateness} summarises the performance of the above-described techniques.
The results tested on the Problem 2 instances as they are originally designed for testing objectives regarding lateness.
SPT and LPT, which are not designed for the objective, as expected achieve poor results and only slightly outperform random plans.
MCTS, however, achieves even better performance than EDD. This showcases the domain-independence strength of MCTS techniques.

It is noteworthy that in all cases the standard deviation is much higher than in previous experiments, which have makespan minimisation as optimisation objectives. This observation together with the worse results for the heuristics show that minimising the maximum lateness is a harder objective than minimising the makespan.

\begin{figure}[H]%
	\centering
	\subfloat[EDD 2.5 opt]{\includegraphics[width=0.45\textwidth]{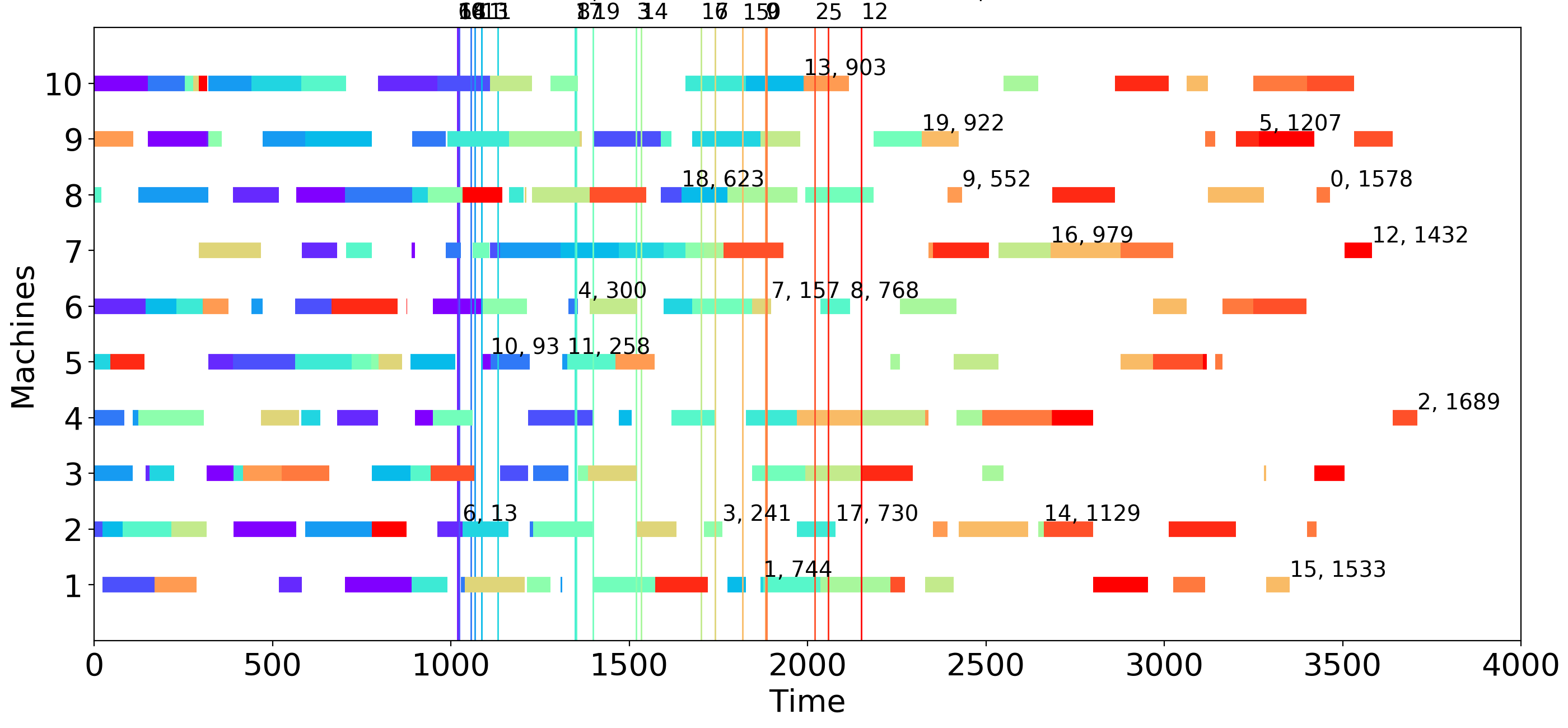}  \label{fig:sl_edf}} 
	\subfloat[Random 4.2 opt]{\includegraphics[width=0.45\textwidth]{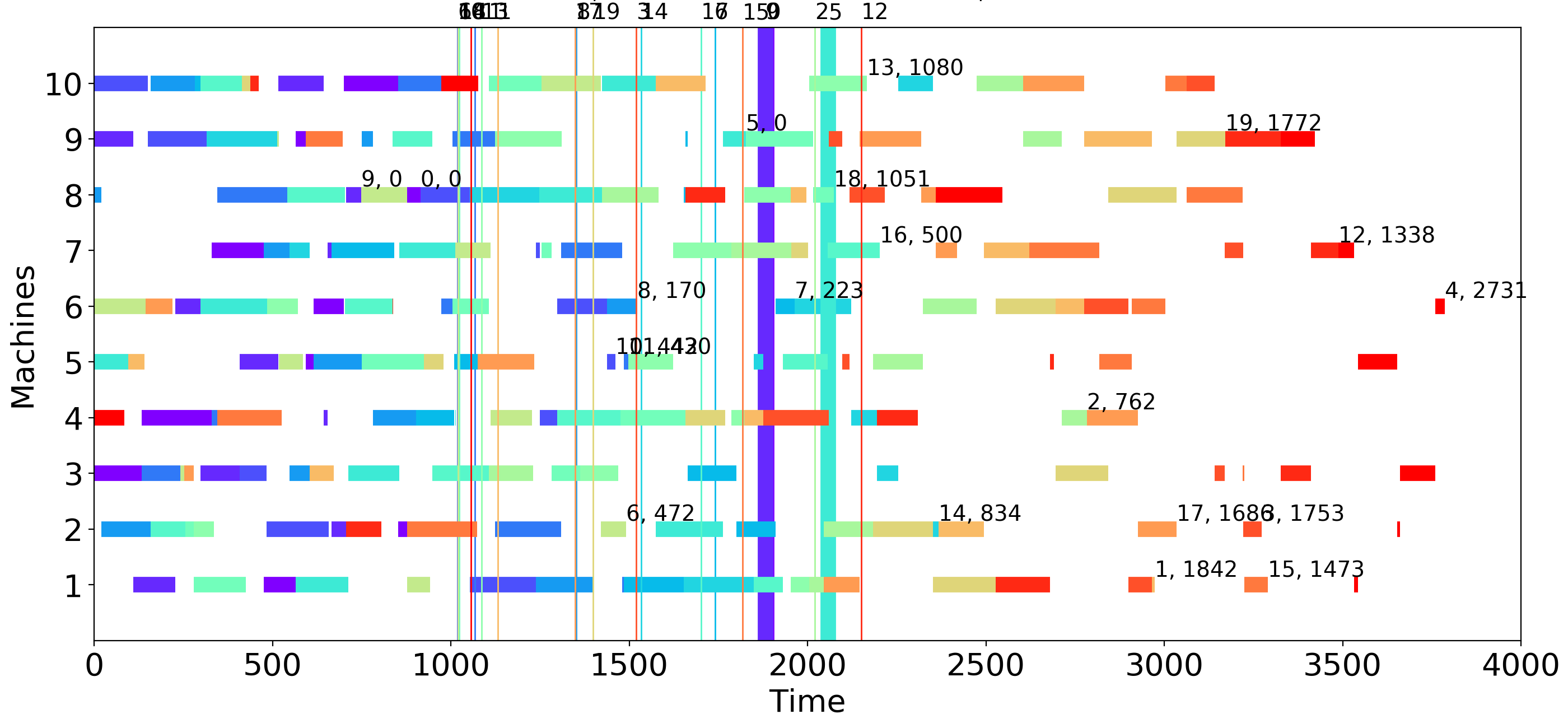}  \label{fig:sl_rand}}\\ 
	\subfloat[Optimal 1 opt]{\includegraphics[width=0.45\textwidth]{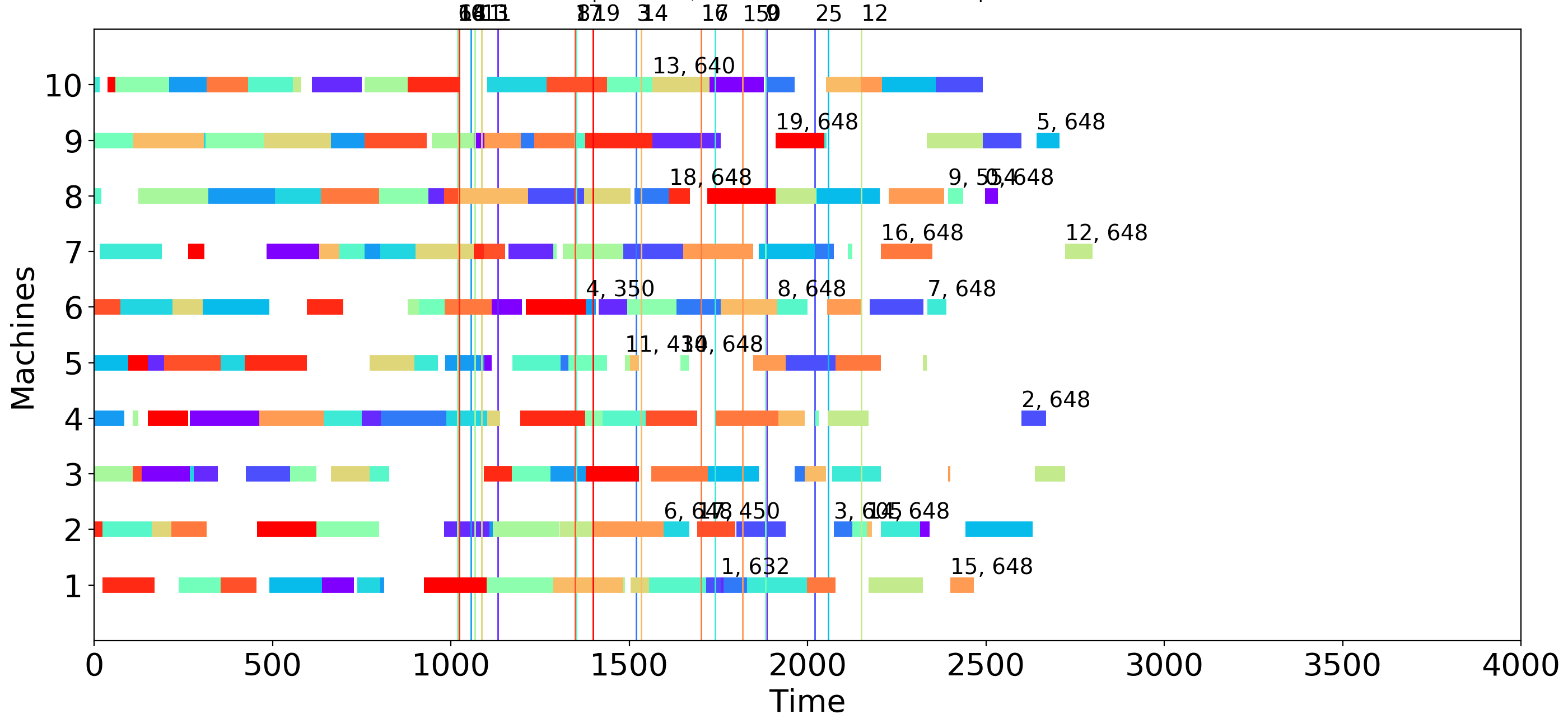}  \label{fig:sl_opt}} 
	\subfloat[MCTS 1.8 opt]{\includegraphics[width=0.45\textwidth]{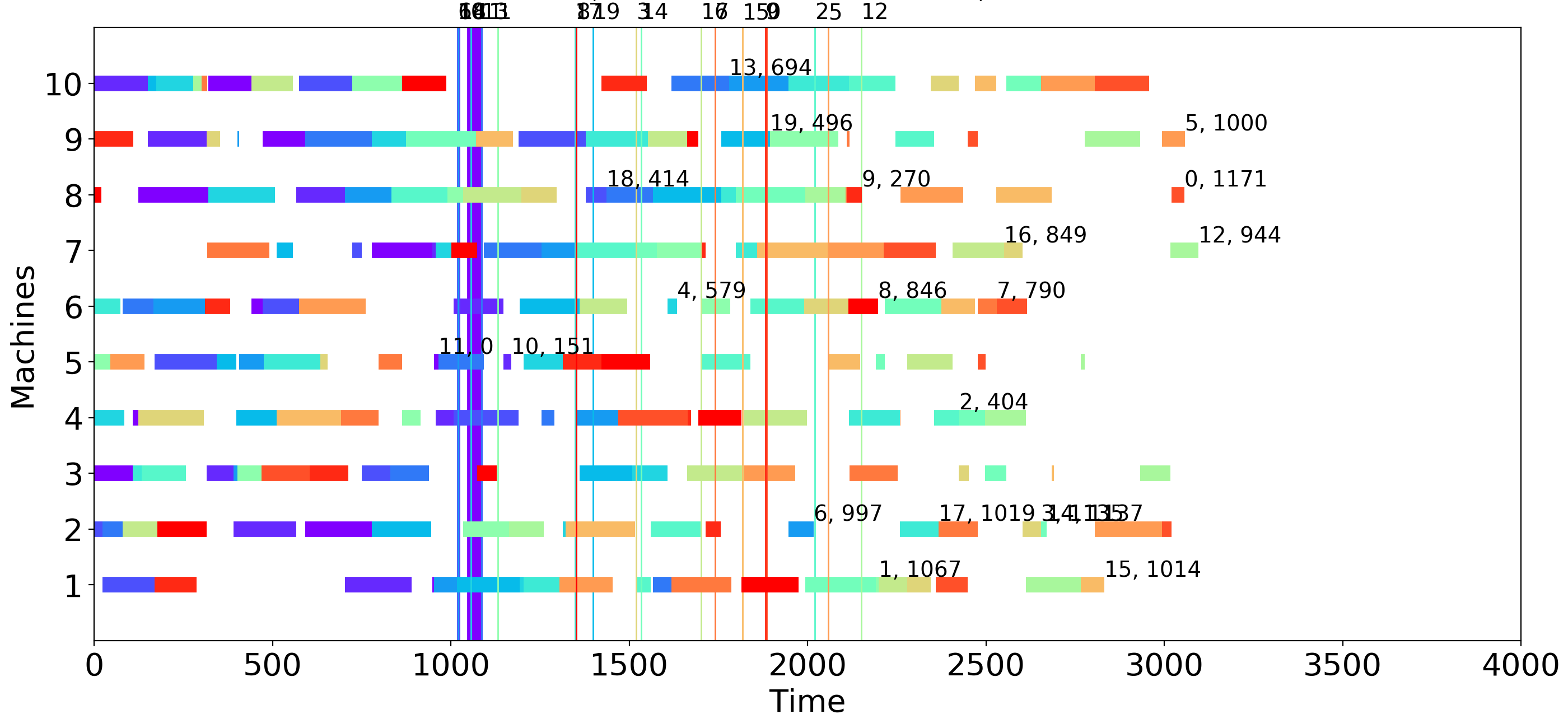}  \label{fig:sl_mcts}}
	\captionsetup{width=.9\linewidth}
	\caption{Resulting schedules of 20 jobs of Problem 2 with minimising maximum lateness as objective. Subfigures (a) to (c) are for reference and (d) is the result after planning with enhanced MCTS. The vertical coloured lines represent the deadlines of the jobs and are visualised in bold if the deadline is not violated. A full description of the jobs is given in \cite{wimmenauer2019}.}%
	\label{fig:schedules_lateness}%
\end{figure}

Schedules created by the compared techniques are in Figure \ref{fig:schedules_lateness}. 
For clear visualisation, as previously, the small JSP instance from  \cite{wimmenauer2019} is used. 
Compared to previous schedules in Figure \ref{fig:schedules}, the ones here are visually much less dense.
This corresponds to the fact that the makespan is not in the optimisation objective anymore.

\begin{figure}[H]%
	\centering
	\subfloat[EDD-Heuristic.]{\begin{tikzpicture}
\pgfplotsset{every axis y label/.style={
		at={(0,0.5)},
		xshift=-5pt,
		rotate=90}}
\pgfplotsset{every axis x label/.style={
		at={(0.5,0)},
		below,
		yshift=0pt}}
\pgfplotsset{ticks=none}
\begin{axis}[enlargelimits=false, axis on top, axis equal image, width=0.58\textwidth,ylabel={\tiny Time},xlabel={\tiny Time},axis line style={draw=none}, xtick=\empty, ytick=\empty]
\addplot graphics [xmin=0,xmax=128,ymin=0,ymax=96] {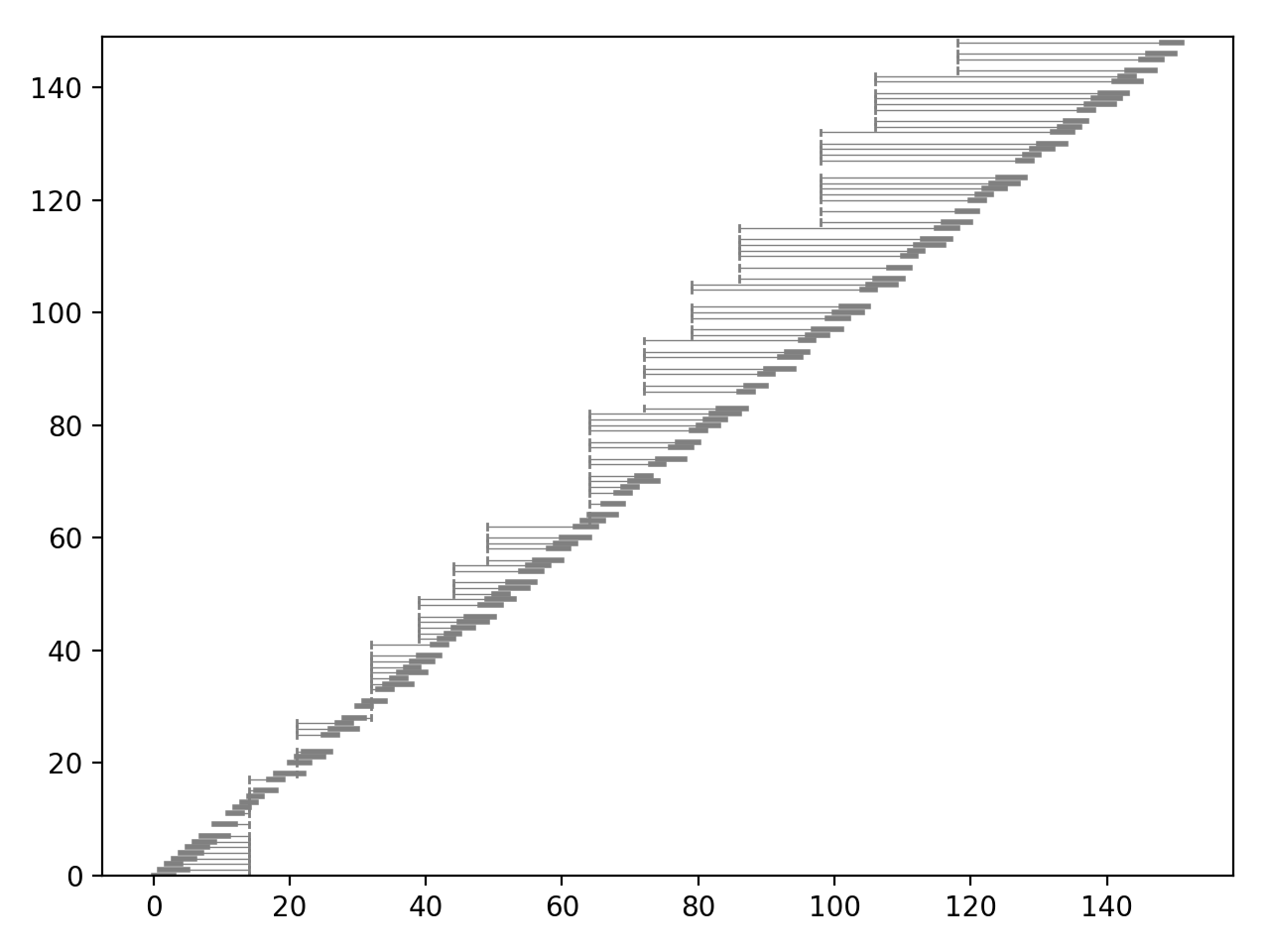};

\end{axis}
\end{tikzpicture}

 	   \label{fig:m_edd}} 
	\subfloat[MCTS with $L_{\max}$ as evaluation metric.]{\begin{tikzpicture}
\pgfplotsset{every axis y label/.style={
		at={(0,0.5)},
		xshift=-5pt,
		rotate=90}}
\pgfplotsset{every axis x label/.style={
		at={(0.5,0)},
		below,
		yshift=0pt}}
\pgfplotsset{ticks=none}
\begin{axis}[enlargelimits=false, axis on top, axis equal image, width=0.58\textwidth,ylabel={\tiny Time} ,xlabel={\tiny Time},axis line style={draw=none}, xtick=\empty, ytick=\empty]
\addplot graphics [xmin=0,xmax=128,ymin=0,ymax=96] {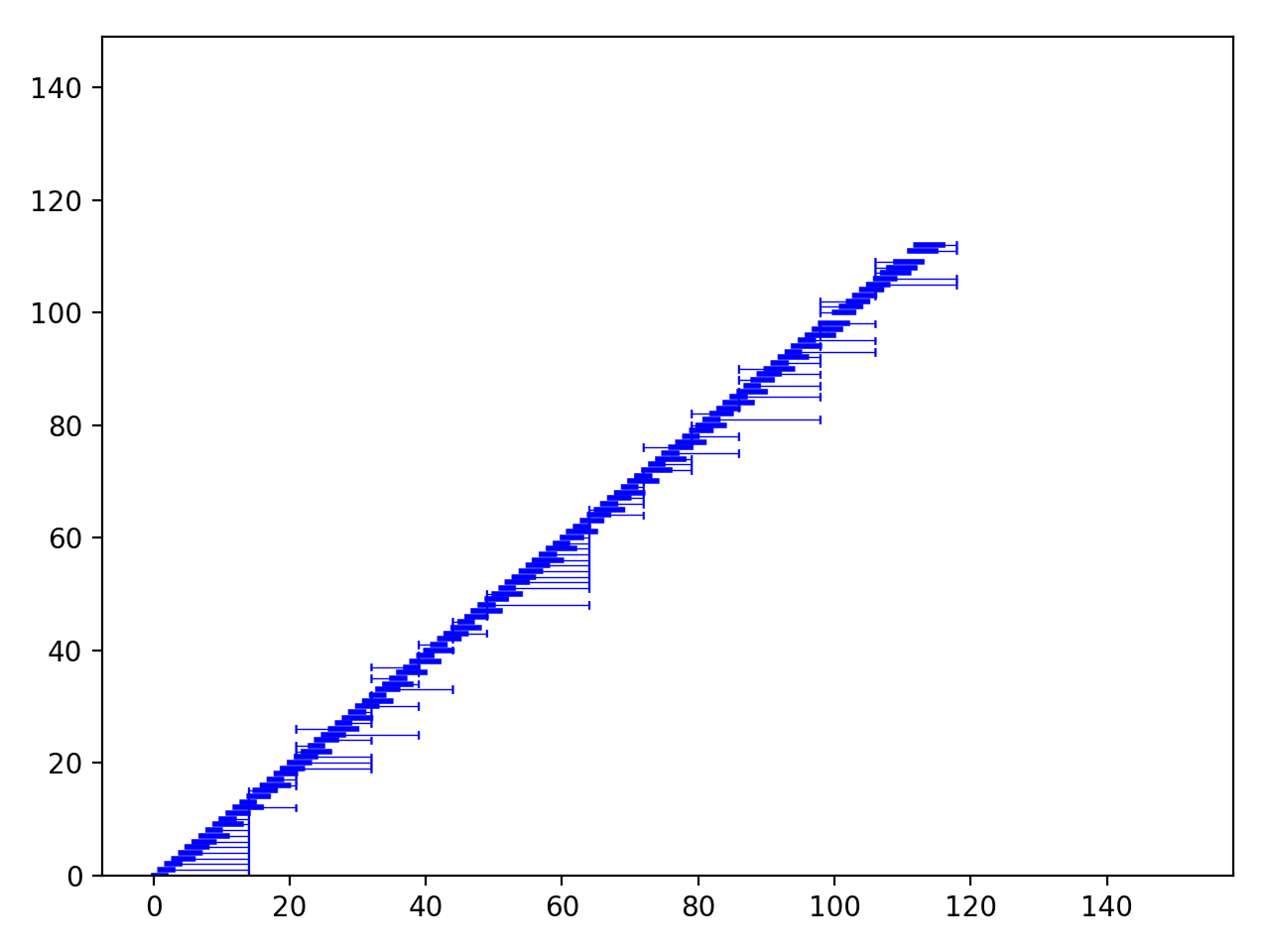};

\end{axis}
\end{tikzpicture}

 	   \label{fig:m_mcs}} 
	
	\captionsetup{width=.9\linewidth}
	\caption{Minimising maximum lateness on instance of Problem 1 with 100 jobs and randomly selected deadlines. Bold horizontal lines depict scheduled jobs, vertical lines depict deadlines and thin horizontal lines are visual support connecting jobs to their deadlines. Deadline is violated if it lies left of the corresponding job. }%
	\label{fig:minLate}%
\end{figure}

As an additional qualitative analysis, the output of EDD and MCTS are shown in Figure \ref{fig:minLate}. 
This is based on Problem 1, which can be more clearly visualised. 
As Figure \ref{fig:m_edd} shows, the delay when using EDD accumulates over time as a result of the heuristic following the single rule of first planning jobs with earliest deadline.
In contrast, Figure \ref{fig:m_mcs} illustrates a much more compact resulting schedule from MCTS. 
The difference shows that MCTS adapts itself to minimise maximum lateness, as the new goal is incorporated into the evaluation metric.
It is unsurprising that the performance exceeds that of EDD, which seeks to indirectly achieve the goal with the heuristically chosen rule of earliest-due-date-first.

%%%%%%%%%%%%%%%%%%%%%%%%%%%%%%%%%%%%%%%%%%%%%%%%%%%%%%%%%%%%%%%%%%%%%%%%%%%%%%%%%

\subsection{Industrial Case Study}  \label{sec:expCaseStudy}

This section presents the experimental results in the context of the industrial case study (Problem 3) described in Subsection \ref{subsec:industrial_case}. Firstly, discussed is a similar case to Subsection \ref{sec:expObjective}, which involves modifying the optimisation objective. Secondly, a qualitative analysis is conducted to illustrate that external resources are handled with the search. The search technique used is H-MCTS with integrated abstraction and all search enhancements. Noteworthy to mention is that the resulting planning system involving H-MCTS  has been implemented and been operating in a production plant's controller since December 2018.

\subsubsection{Minimising Mixed Objective}
In this case, the goal is to minimise the makespan as well as the maximum lateness. 
The plan evaluation in the search is therefore modified as $score = \frac{\max_{j\in J}(c_j)}{|J|}+L_{\max}$, where $c_j$ is the completion time of job $j$, $L_{\max}$ is the maximum lateness.
The denominator in the first term serves as a normalisation factor, since the makespan grows proportionally with the number of jobs.

As there are two components in the new objective,
it is difficult to find an existing heuristic specifically for this purpose.
Furthermore, due to the encapsulated scheduling heuristic, there is no optimum or bound available either.
The only possible performance comparison is against the original situation before the implementation of the new planning system.
Table \ref{tab:case_schedule} outlines the performance of the original and new planning system in the production plant on 6 planning situations, as well as the relative improvement of the new system.

\begin{table}[H]
	\definecolor{highlightColor}{gray}{0.9}
	\centering
%	\resizebox{\textwidth}{!}{
		\begin{tabular}{@{}rcccccccc@{}}
			\toprule
			$|J|=$ & 1139 & 804 & 1037 & 1072 & 806 & 1121 & Avg. & Std.dev. \\
			\midrule
			\multicolumn{1}{l}{Original}\\
			$c_{\max}$ &1493	& 1039	& 1348	& 1382	& 1043	& 1494	& &\\%1299.8	& 208.8\\
			$L_{\max}$ & 49		& 51		& 66		& 69		& 50		& 78		& &\\%60.50	& 12.18\\
			$score$ & 	50.31	& 52.29	& 67.30	& 70.29	& 51.29	& 79.33	& &\\%61.80	& 12.19\\
			\multicolumn{1}{l}{Planned}\\
			$c_{\max}$ &1371	& 937		& 1229	& 1218	& 934		& 1289	& &\\%1163	& 184.4\\
			$L_{\max}$ & 47		& 51		& 65		& 67		& 49		& 78		& &\\%59.50	& 12.39\\
			$score$ & 48.20	& 52.17	& 66.19	& 68.14	& 50.16	& 79.15	&&\\% 60.67	& 12.38\\
			\multicolumn{1}{l}{Improvement}\\
			$c_{\max}$ &8.2\%	& 9.8\%	& 8.8\%	& 11.9\%	& 10.5\%	& 13.7\%	& 10.5\%	& 20\%\\
			$L_{\max}$ & 	4.1\%	& 0.0\%	& 1.5\%	& 2.9\%	& 2.0\%	& 0.0\%	& 1.7\%	& 1.6\%\\
			$score$ & 4.2\%	& 0.2\%	& 1.7\%	& 3.1\%	& 2.2\%	& 0.2\%	& 1.9\%	& 1.6\%\\
			\bottomrule
		\end{tabular}
%		}  %%% END OF RESIZEBOX
		\captionsetup{width=.9\linewidth}
		\caption{Minimising mixed objective in Problem 3. The makespan, lateness and score are reported for two scenarios. Firstly, labelled as ``Original", for  schedules generated without the search approach. Secondly, labelled as ``Planned", for the approach presented here}
		\label{tab:case_schedule}
\end{table}

Over the 6 plans, the maximum lateness improved by 1.7\% and the makespan by 10.5\%.
Even after modifying the relative weights of lateness and makespan in the evaluation score, no further improvement in the makespan can be achieved. 
This implies that it may be generally harder to shorten the maximum lateness.

Finally, the result of the integrated abstraction is a hierarchy where the most abstract level joins options based on deadline similarity. More concrete hierarchy levels join options on process similarity. This hierarchy of abstraction is explainable by the large impact of the deadlines to the score. 

\subsubsection{Handling External Resources}
The industrial case study involves several additional constraints as described in Subsection \ref{subsec:industrial_case}, namely the presence of racks, carriers, and the loading area, which all need to be modelled when searching for plans.
This experiment zooms into a specific plan to showcase how these additional constraints are handled.

Table \ref{tab:externals} shows part of the log of a created plan.
The highlighted rows all require rack type 14. 
There are, however, only three racks of this type.
Job 1040 on carrier-rack pair 77-14 is planned at time step 527 and completes at 560. 
As soon as this job finishes and its carrier-rack pair becomes available at time step 560, job 1043 is scheduled at time step 561, reusing the carrier-rack pair 77-14.
The same kind of carrier-rack pair re-usage also occur with 78-14 and 79-14.
The jobs with the same carrier-rack pairs are colour-coded.
It can be seen that the re-usage happens directly after the require carrier-rack pair is free.

Furthermore, the effects of the loading area, which is modelled as a polling system are visible in the plan.
It is always the case that two jobs are planned to start simultaneously.

It appears that the heuristic combines pairs of jobs with same process. Jobs 1040 and 1041, both having process 3, are planned at different time but completed at the same time. The same holds for jobs 1042 and 967. 
Prior to being processed, job 1060 performed rack change. 
This job has 40 steps between planned start and completion time even though process 2 has processing time of 30 steps. 
A rack change takes 10 steps, which explains the long time in job is active. 

\begin{table}[H]
	\definecolor{highlightColor}{gray}{0.9}
	\definecolor{highlightColor1}{gray}{0.8}
	\definecolor{highlightColor2}{gray}{0.7}
	\centering
	\begin{tabular}{@{}ccccccc@{}}
		\toprule
		Job Id  & Process 	& Carrier & Rack & $d_j$ & $s^p_j$ & $c_j$ \\
		\midrule
		\multicolumn{7}{c}{$\vdots$}\\
		\rowcolor{highlightColor}
		1040 	& 3				& 77 		& 14 			 & 	2500	& 527 		&  560 \\ %32+1
		837 	& 4				& 24 		& 232 			 & 	1450	& 527 		&  564 \\  %35 +2
		\rowcolor{highlightColor1}
		1041 	& 3				& 78 		& 14 			 & 	2500	& 528 		&  560 \\%32
		\rowcolor{highlightColor2}
		1042 	& 3				& 79 		& 14 			 & 	2500	& 528 		&  561 \\%32 +1
		967 	& 3				& 8 		& 154 			 & 	 900	& 529 		&  561 \\ %32
		838 	& 4				& 25 		& 232 			 & 	1450	& 529 		&  564 \\%35 
		98 	& 2				& 57 		& 78 			 & 	550	     & 530 		&  570 \\%30 + 10
		\multicolumn{7}{c}{$\vdots$}\\
		972 	& 3				& 8 		& 154 			 & 	 900	& 561 		&  593 \\ %32
		\rowcolor{highlightColor}
		1043 	& 3				& 77 		& 14 			 & 	2500	& 561 		&  593 \\ %32
		\rowcolor{highlightColor1}
		1044 	& 3				& 78 		& 14 			 & 	2500	& 562 		&  594 \\%32
		\rowcolor{highlightColor2}
		1045 	& 3				& 79 		& 14 			 & 	2500	& 562 		&  594 \\%32
		973 	& 3				& 8 		& 154 			 & 	 900	& 563 		&  595 \\ %32
		\multicolumn{7}{c}{$\vdots$}\\
		\bottomrule
	\end{tabular}
	\caption{An excerpt of plan showing the planned jobs with planned external resources and times of deadline, planned start and completion time. The three highlighted pairs of rows show the re-usage of external resources upon completion of previous jobs.}
	\label{tab:externals}
\end{table}

%%%%%%%%%%%%%%%%%%%%%%%%%%%%%%%%%%%%%%%%%%%%%%%%%%%%%%%%%%%%%%%%%%%%%%%%%%%%%%%%%
%%%   Conclusions
%%%%%%%%%%%%%%%%%%%%%%%%%%%%%%%%%%%%%%%%%%%%%%%%%%%%%%%%%%%%%%%%%%%%%%%%%%%%%%%%%

\section{Conclusions}
\label{sec:conclusions}
The article considered the case of job shop problems where the underlying scheduling heuristic is unknown and has to be considered as blackbox. As the performance of such a scheduling heuristic often depends on the ordering of the jobs, it is therefore a search problem. This article has  proposed a technique that uses Monte-Carlo Tree Search to find input sequences to the scheduling heuristic such that its performance is maximized. To cope with large solutions space in planning scenarios, a Hierarchical Monte-Carlo Tree Search (H-MCTS) is proposed based on abstraction of jobs.  A feed forward network  has been used to model for H-MCTS the behaviour of the scheduling heuristic  as it had the best trade-off in terms of speed and accuracy. 

The experiments revealed that on synthetic and real-life problems,  H-MCTS with integrated abstraction significantly outperforms pure heuristic-based techniques as well as other Monte-Carlo search variants. Performance could be further improved by including search enhancements originating from the MCTS literature. 
The experimental results furthermore showed that, by modifying the evaluation metric in H-MCTS, it is possible to achieve other optimization objectives than what the scheduling heuristics are designed for. This observation has been validated also in real-life cases, as H-MCTS has been implemented in a production plant's controller. 
 
There are several directions of future research. First, the large number of jobs leads to long roll-outs and evaluations of long plans, both of which are computationally expensive. To overcome this problem, more research could be conducted on the challenging task of evaluation abstraction.  This would allow the hierarchical search to  evaluate directly shorter sequences of abstract states without expanding the hierarchical search down to the primitive states. With this hierarchical evaluation, expansion of the hierarchy would only happen in promising regions of the search space.

Second, this article  demonstrated promising results on integrating automatic abstraction into search. However, on the uniformly random generated Demirkol instances, the integrated abstraction did not show performance improvement due to the lack of structure in the jobs.  Further research could extend the current work by handling problems with less structure. 

Additionally, further improving the performance of the MCTS for the planning problem at hand could result in better plans.  Currently, the roll-outs generate job sequences at random. This can be extended by learning better roll-out strategies for example by using N-grams in the roll-out \cite{tak2012n}.

\subsubsection*{Acknowledgements}
The authors would like to thank Michael Leer and Andreas Schuhmacher for their help on the industrial case study.

%%%%%%%%%%%%%%%%%%%%%%%%%%%%%%%%%%%%%%%%%%%%%%%%%%%%%%%%%%%%%%%%%%%%%%%%%%%%%%%%%
%%%   BIBLIOGRAPHY
%%%%%%%%%%%%%%%%%%%%%%%%%%%%%%%%%%%%%%%%%%%%%%%%%%%%%%%%%%%%%%%%%%%%%%%%%%%%%%%%%

\bibliographystyle{plain}
\bibliography{mcts-for-blackbox-scheduling}

\end{document}